\documentclass[sigconf,nonacm]{acmart}
\renewcommand\footnotetextcopyrightpermission[1]{} 

\usepackage{latexsym}

\usepackage{multirow}
\usepackage{svg}
\usepackage{paralist}
\usepackage{float}

\usepackage[bibliography=separate]{apxproof} 
\acmConference[JCDL'24]{Joint Conference on Digital Libraries}{December 2024}{Hong Kong}

\usepackage{amsmath}
\usepackage{amsfonts}

\usepackage{booktabs}
\usepackage{array}

\title{Modular Multimodal Machine Learning for Extraction
of~Theorems~and~Proofs in Long Scientific Documents (Extended~Version)}


\begin{document}


\author{Shrey Mishra}
\affiliation{%
	\institution{DI ENS, ENS, CNRS, PSL University, Inria}
	\city{Paris}
	\country{France}
}
\email{shrey.mishra@ens.psl.eu}

\author{Antoine Gauquier}
\affiliation{%
	\institution{DI ENS, ENS, CNRS, PSL University, Inria}
	\city{Paris}
	\country{France}
}
\email{antoine.gauquier@ens.psl.eu}

\author{Pierre Senellart}
\affiliation{%
	\institution{DI ENS, ENS, CNRS, PSL University, Inria \& IUF}
	\city{Paris}
	\country{France}
}
\email{pierre@senellart.com}

\begin{abstract}
We address the extraction of mathematical statements and their proofs
from scholarly PDF articles as a multimodal classification problem,
utilizing text, font features, and bitmap image renderings of PDFs as
distinct modalities. We propose a modular sequential multimodal machine
learning approach specifically designed for extracting theorem-like
environments and proofs. This is based on a cross-modal attention
mechanism to generate multimodal paragraph embeddings, which are then fed
into our novel multimodal sliding window transformer architecture to
capture sequential information across paragraphs. Our document AI
methodology stands out as it eliminates the need for OCR preprocessing,
\LaTeX{} sources during inference, or custom pre-training on specialized losses to understand cross-modality relationships. Unlike many conventional approaches that operate at a single-page level, ours can be directly applied to multi-page PDFs and seamlessly handles the page breaks often found in lengthy scientific mathematical documents.
Our approach demonstrates performance improvements obtained by transitioning from unimodality to multimodality, and finally by incorporating sequential modeling over paragraphs.
\end{abstract}

\maketitle

\section{Introduction}
\paragraph*{Context.}
Scholarly articles in mathematical fields
typically include theorems (and other
theorem-like environments) along with their proofs. This paper builds upon our previous work
\cite{mishra2024first}, which aimed to transform scientific literature from a collection of PDF
articles into an open knowledge base (KB) centered around theorems.

The objective of \cite{mishra2024first} was to enable new ways of exploring mathematical
results, such as searching for all theorems that depend on a specific result or identifying all
proofs that include a particular feature.

For example, such a knowledge base would allow the following:
\begin{enumerate}
	\item \textbf{Navigating through the scientific literature:}
	      Currently, the only way to navigate through the scientific
	      literature is through search engines such as Google or Google
	      Scholar that index the full-text of papers, or by navigating
	      through citation links. These approaches do not allow indexing of
	      individual mathematical results, which is the main object of
	      interest of mathematicians and theoretical computer scientists.
	      With a KB of scientific results, one would be able to find, e.g.,
	      all NP-hardness results involving the vertex cover problem (and not
	      just all papers that contain both the terms ``NP-hard'' and
	      ``vertex cover'').
	\item \textbf{Identifying the impact of errors in theorems:} Another
	      useful application of such a knowledge base is to determine which
	      theorems are used in the proof of another theorem. This would be of
	      tremendous use, for instance, to determine which results become
	      invalidated or need to be revisited when one of the theorems they
	      depend on is shown to be false.
\end{enumerate}

In \cite{mishra2024first}, we outlined the comprehensive scope of our project and presented preliminary evaluations on two fundamental subtasks:
\begin{compactenum}[(i)]
	\item extraction of information pertaining to proofs and theorems;
	\item linkage of mathematical results across various papers.
\end{compactenum}
In this paper, we concentrate primarily on the extraction aspect of the pipeline introduced in \cite{mishra2024first}. We conduct an in-depth
exploration of diverse multimodal methodologies and assess the impact of modeling long-term paragraph sequences.
This is particularly advantageous for the identification of mathematical results, as it utilizes the contextual information
surrounding the paragraphs covering length proofs.

\paragraph*{Problem definition.}
As a first step towards this ambitious goal of building a knowledge base of mathematical results, it is necessary to develop information
extraction methods that automatically identify theorem-like environments
and proofs in PDF scientific articles.

A human being would typically be able to
perform this task by relying on the formatting of the text, on specific
keywords identifying the environments, and on other visual clues:
including keywords such
as ``Theorem'' or ``Proof'' in bold or italics, the fact that an entire
block of text might be in italics, the comparatively high proportion of
mathematical characters, the presence of a QED symbol at the
end of a proof, etc. However, precise formatting depends on document
formats; a classifier that would only use such kinds of hard-coded rules
does not generalize well for arbitrary formats, or for proofs that span
multiple paragraphs.


To clarify, in the whole of this paper we use \emph{theorem} in the same
sense as it is used in \LaTeX{} (say, by the \verb|\newtheorem| command):
a theorem-like environment is a structured statement, possibly numbered, formatted in a
specific way and used to represent a formal (usually mathematical)
statement: it can be a theorem, a lemma, a proposition, etc., but also a
definition, a formal remark or an example. By \emph{theorem} we mean any statement of this kind. By \emph{proof} we mean
what would typically be rendered in \LaTeX{} in a \texttt{proof}
environment: a proof or proof sketch of a result.

We propose to approach the theorem--proof identification problem by
designing
an approach based on multimodal machine learning that classifies each
paragraph of an article into \emph{basic}, \emph{theorem}, and
\emph{proof} labels, based on the
scientific language, on typographical information, and on visual
rendering of PDF documents.
Additionally, we take into account information about the
\emph{sequence} of paragraph blocks, normalised spatial coordinates and page numbers along with page breaks, to exploit the fact that the label of a paragraph heavily relies on that of the preceding (and possibly following) ones.

\begin{figure*}
	\includegraphics[width=\linewidth]{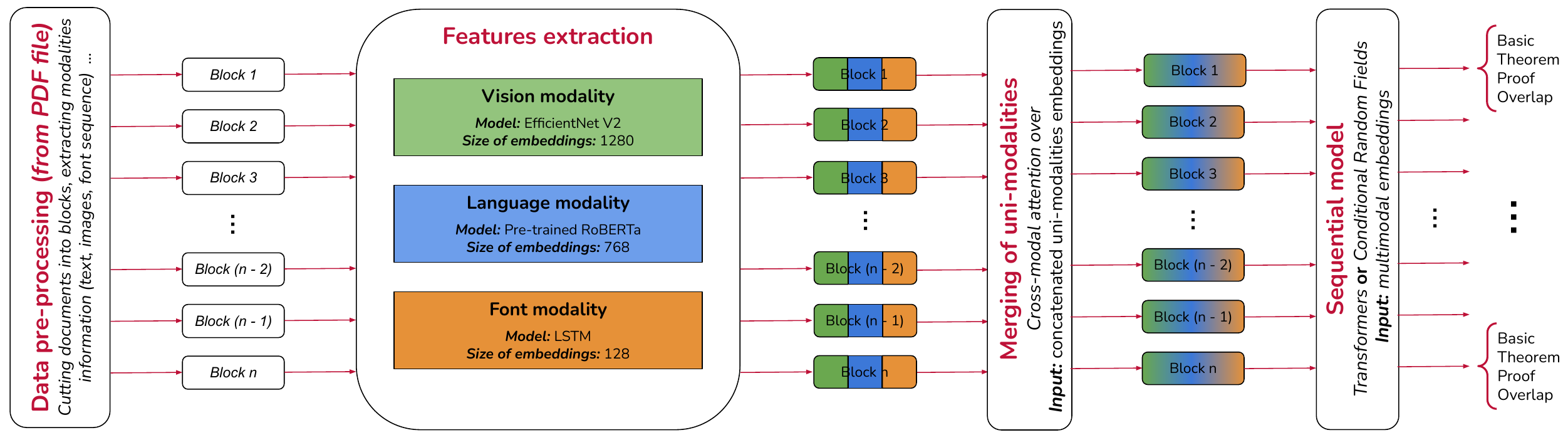}
	\caption{Overall model Inference pipeline}
	\label{fig:pipeline}
\end{figure*}

\paragraph*{Methodology and contributions.}
To design a multimodal approach to the theorem--proof identification problem,
we take inspiration from how a human being would solve the task, i.e., with the help of:
\begin{enumerate}
	\item Understanding of the scientific vocabulary and how mathematical
	      writing is organized: it might be possible to recognize a proof or a
	      theorem by the presence of phrases as “We conclude by” or “Assume
	      \dots Then \dots”.
	\item Visual features such as symbols and the use of bold or italic
	      fonts: some document classes, for instance, format the content of a
	      theorem all in italics and end all proofs with a QED symbol.
	\item Use of different font types and sizes in order within paragraphs: starting a paragraph with a word in bold or in italics.
	\item Sequential organization of blocks within a document: For example,
	      if we know the label of both previous and next paragraphs are
	      \emph{proof}, it is likely that of the current paragraph is also
	      \emph{proof} (or possibly a \emph{theorem}; \emph{basic} is unlikely);
	      this is even more relevant, if vertical spacing between these blocks is
	      small.
\end{enumerate}

This suggests, respectively, the use of a language model to capture
text-level information; the use of a computer-vision approach to capture
visual features; the use of styling information to capture font-based
information; and the use of a sequential model to capture information
from block sequences. In addition, we want to be able to combine all
these features in a unified multimodal approach.

We provide the following contributions in this paper, summarized in
Figure~\ref{fig:pipeline}:
\begin{inparaenum}[(i)]
	\item Three unimodal (vision, text, font information) models for the
	theorem--proof identification problem relying on modern machine learning
	techniques (CNNs, transformers, LSTMs) with a focus on reasonably
	efficient models as opposed to very large ones; note that the text
	modality approach relies on pretraining a language model specific to
	our corpus, which may have applications beyond our task.
	\item A multimodal late fusion model that combines the features of all
	three modalities.
	\item A block sequential approach, based on a
	transformer model, that can be used to improve the performance of any
	unimodal and multimodal model by capturing dependencies between
	blocks.
	\item An experimental evaluation on a dataset of roughly 200k
	English-language papers from
	arXiv, with a separate validation dataset of 3.5k papers (amounting to
	529k paragraph blocks).
\end{inparaenum}

\paragraph*{Outline.}
After discussing related work in Section~\ref{sec:related}, we present in Section~\ref{sec:unimodal} the three unimodal
models. We then discuss in Section~\ref{sec:beyond-unimodal} how to combine
them into a multimodal model, and how to add support for information
about block sequences. We further provide a description of our dataset
in Section~\ref{sec:dataset}.
Experimental results on all unimodal
and multimodal models is presented in in Section~\ref{sec:experiments}.
For brevity, additional materials are provided as supplementary content. Detailed information,
including design choices, architecture diagrams,
confusion matrices for all classes, data pipeline diagrams, explainability, large-scale societal impact, and
other critical aspects of the project, are comprehensively
discussed in the PhD thesis of the first author~\cite{mishra2024multimodal}.
The code, data, and models supporting this paper are accessible at \url{https://github.com/mv96/mm_extraction}.

\section{Related Work}
\label{sec:related}

We discuss now related work about extraction of theorems and proofs from
the scientific literature, and more broadly about document datasets.
We shall discuss further related work relevant to unimodal or multimodal
approaches when discussing individual models.

\paragraph*{Extraction of theorems and proofs}
The theorem--proof extraction problem has received little interest in past
research, though we now discuss two highly relevant works
\cite{DBLP:journals/corr/abs-1908-10993,mishra2021towards}.

\citet{DBLP:journals/corr/abs-1908-10993} proposed the task
of identifying proofs and theorem-like environments from
arXiv\footnote{\url{https://arxiv.org/}} articles
using their HTML rendering via \LaTeX{}ML. Their approach involves
detecting mathematical statements (along with other regions such as
abstract and acknowledgements), introduced as a 50-class classification
problem. They show that there is some link in the contextual
information among paragraphs, which is then exploited by the textual
modality over a BiLSTM-based encoder/decoder approach. This approach has
two major limitations, which make it unsuitable for our
needs:
\begin{inparaenum}[(i)]
	\item Their approach does not operate on raw PDFs but on HTML
	renderings, which makes it only applicable when \LaTeX{} source code is
	available.\footnote{Note that in such settings, extracting theorems
		and proofs from the
		source, as we do in Section~\ref{sec:dataset} to build a labeled
		dataset, seems
		a better alternative.}
	\item Their approach is only evaluated on the first logical
	paragraph within a marked-up environment belonging to the label
	set (e.g., only the first paragraph of every proof), which makes the task much simpler, since the first word is
	highly indicative of the label in most cases; in contrast, we aim
	at differentiating such environments from regular text, and we
	aim at classifying all paragraphs within an environment, not just
	the first one.
\end{inparaenum}
In addition, note that accessing the
dataset of~\cite{DBLP:journals/corr/abs-1908-10993}
requires signing an NDA\footnote{\url{https://sigmathling.kwarc.info/resources/arxmliv-statements-082018/}} whose terms prevent free use for research.

In our prior work~\cite{mishra2021towards}, we
built a proof of concept system evaluating
various unimodal approaches based on different evaluation metrics using
NLP, computer vision, and a mix of heuristics (detection of specific
keywords) and
font-based information to
identify mathematical regions of interest. The problem was posed as a
3-class classification problem operated on text \emph{lines} extracted from raw
PDFs obtained using
\textsf{pdfalto}\footnote{\label{ft:pdfalto}\url{https://github.com/kermitt2/pdfalto}}.
This work had some important limitations:
\begin{inparaenum}[(i)]
	\item Text lines do not usually contain entire sentences and offer little context, which means they are hard to classify.
	\item The computer vision approach was framed as an object detection challenge, utilizing an Intersection Over Union (IOU) based metric, poorly suited for identifying text blocks, as alterations in the threshold can affect the detection score. Additionally, this discrepancy complicates the integration of the computer vision approach with other modalities that function at the text line level.
	\item The third modality, focusing on heuristics and font features, primarily utilized hand-crafted characteristics, including checks for whether the first word is bold or italic.
	\item Moreover, \citet{mishra2021towards} do not present a consistent method for comparing the three modalities, as each modality relies on a distinct segment of information. In our study, we strive to establish a standardized approach for evaluating the performance of various modalities, as well as combining them together.
\end{inparaenum}
This study builds upon~\cite{mishra2021towards}, addressing and overcoming the four primary limitations
identified therein. We introduce a modular, multimodal framework that facilitates a consistent methodology for the
comparison and integration of the three modalities, eliminating the necessity for \LaTeX{} source files during inference.

\paragraph*{Document datasets.}
In this research, we utilize \textsf{Grobid} \cite{grobid} to extract textual content and
bounding boxes from paragraph-rendered bitmap images, a foundational step
for training our models. This work shares similarities with the
objectives outlined in the \textsf{Publaynet} study \cite{zhong2019publaynet},
particularly in our focus on identifying ``Proofs'' and ``Theorems'' within
scholarly articles. Unlike \textsf{Publaynet}, which categorizes document sections
into Text, Tables, Figures, etc., at the document level, our study
extends to the analysis of academic writings, leveraging \LaTeX{} sources
from arXiv submissions for ground truth generation, in contrast to
\textsf{Publaynet}'s use of the National Library of Medicine
(NLM)\footnote{\url{https://dtd.nlm.nih.gov/}} schema for journal articles.

At its core, \textsf{Grobid} covers a broad array of document segments,
including lists, figures, titles, and bibliographic entries, akin to the
scope of \textsf{Publaynet} \cite{zhong2019publaynet}. It distinguishes
itself by semantically parsing texts into sentences and paragraphs and
accurately identifying block coordinates, thus supporting a text-based
modality in our analysis. Notably, \textsf{Grobid} preserves mathematical content
within textual segments, a feature not prioritized by Publaynet, which
omits certain XML tree nodes like tex-math and disp formula (as stated by
the PMCOA XML on Page 2 of \cite{zhong2019publaynet}). This distinction underscores the relevance of our approach to the specific requirements of our study and its broader goals. While Publaynet prioritizes visual classification among diverse labels such as Text, Figures, and Tables, our analysis, as evidenced in Table \ref{tab:crf_results}, highlights the indispensable role of textual modalities in distinguishing proofs across paragraphs, underscoring the multimodal nature of our challenge where textual analysis is paramount.

The debut of \textsf{Docbank} \cite{li2020docbank} marks a notable
advancement beyond \textsf{Publaynet} \cite{zhong2019publaynet}, offering
a comprehensive dataset of 500K document images tailored for training and
testing applications. Unlike Publaynet's emphasis on the medical field,
\textsf{Docbank} encompasses a wider array of academic areas, including
Physics, Mathematics, and Computer Science. This diversity introduces a
rich variety of mathematical formulas to the dataset. \textsf{Docbank}'s
distinctive feature is its dual-level annotations -- both token and
segment -- rendering it highly applicable for a broad spectrum of tasks
in computer vision and natural language processing. Despite its potential
for aiding proof identification projects, \textsf{Docbank}'s broad
annotation scope, encompassing author names, abstracts, titles,
equations, and paragraphs, may dilute its applicability for our focused
research on identifying proofs and theorems within texts. A limitation
arises from \textsf{Docbank}'s lack of specific labels for proofs or
theorems, complicating its use for our problem, given our dataset's focus on documents that definitively contain proofs and theorems.

A noteworthy feature of \textsf{Docbank} is its sourcing of documents
from arXiv, associating each with an arXiv ID. This linkage permits
access to the \LaTeX{} sources of the papers, enabling the application of
our preprocessing script for ground truth annotation (proofs and
theorems) within the \textsf{Docbank} dataset, thereby broadening its utility. The
adoption of the \verb|\begin| command for annotations by \textsf{Docbank}'s authors parallels the methodology utilized in our research for marking structural segments in scientific documents, illustrating a shared approach in identifying and analyzing document components like proofs and theorems.

\textsf{Doclaynet} \cite{pfitzmann2022doclaynet}, paralleling the efforts
of \textsf{Docbank} \cite{li2020docbank} and \textsf{Publaynet}
\cite{zhong2019publaynet}, targets layout detection in documents with a
focused dataset of 80K instances. This dataset extends beyond scientific
articles to include a broader range of paper layouts, aiming to achieve
the detection precision of models like \textsf{FASTRCNN}
\cite{ren2015faster} and \textsf{YOLOv5}
\cite{glenn_jocher_2020_4154370}. A notable challenge identified in
\textsf{Doclaynet} is the presence of overlapping labels, where blocks
share intersecting labels, a complexity also acknowledged in our
research. To mitigate this, both studies prioritize the analysis of
non-overlapping blocks for evaluation. Interestingly, our validation
dataset, encompassing approximately 80K images, aligns closely in scale
with \textsf{Docbank}'s (around 50K) and exceeds that of
\textsf{Doclaynet} (around 6K images), providing a substantial basis for
our task. Our findings further reveal that a relatively modest collection
of a few thousand PDFs
suffices to enhance performance on the validation data, underscoring the efficiency of our unimodal approaches.

\section{Unimodal Models}
\label{sec:unimodal}
We now present the methodology of our three unimodal models: a
pretrained transformer (\textsf{RoBERTa}-based) language model for text
extracted for each paragraph of the PDF; an
\textsf{EfficientNetv2M}~\cite{tan2019efficientnet} CNN for vision on the
bitmap rendering of each PDF paragraph; and an LSTM model
trained on font information sequences within each paragraph.
For a technical reason explained in Section~\ref{sec:dataset}, the
problem is formulated as a four-class classification: in addition to the
three target \emph{basic
text}, \emph{theorem}, \emph{proof}, we employ a reject \emph{overlap} class.

\label{sec:methodology}
\subsection{Text Modality}
\label{sec:text}

\paragraph*{Pretraining language models.}


\begin{toappendix}
\subsection{Text Modality}
We investigate the difference in base
vocabulary of different language models: Figure~\ref{fig:vocab_diff_cf}
compares the overlap of vocabulary between various language models -- the
one we are proposing (\texttt{trained\_tokenizer} in the figure)
has a maximum of 33\% overlap with
others, including \textsf{SciBERT} which is trained on scientific text, suggesting
the relevance of a pretrained model with vocabulary specific to our corpus.

\begin{figure}[h]
\centering\includesvg[width=.8\linewidth]{images/vocabulary_overlap.svg}
\caption{Vocabulary overlap among popular language models (BERT,
DistilBERT, \textsf{SciBERT}, in cased or uncased variants) and our pretrained
model (labeled as \texttt{trained\_tokenizer} here)}
\label{fig:vocab_diff_cf}
\end{figure}

In our pretraining, we used 11~GB of pretraining text data (196\,846
scientific articles, see Section~\ref{sec:dataset}), trained over 11
epochs. We used the LAMB optimizer~\cite{you2019large} and produced the
results using a total batch size of 256 across 4 NVIDIA A100 GPUs
with a distributed mirrored strategy and an initial learning rate of $2\times
10^{-5}$. The total pretraining time was 176 hours.

Beyond vocabulary, the nature and size of pretraining data critically
influence a language model's performance and its speed of generalization.
Research, including the Chinchilla study~\cite{hoffmann2022training},
highlights the significant impact of data size, introducing the
``Chinchilla scaling laws''. For example, Chinchilla (70B) outperformed Gopher (280B) by 7\% by quadrupling its training data, demonstrating these principles even beyond a trillion tokens, as seen with the LLAMA \cite{touvron2023llama} model (7B) surpassing GPT-3 (175B \cite{brown2020language}).

In this paper, we choose not to utilize several billion-parameter models for the reasons outlined below:
\begin{itemize}
    \item \textbf{Focus on integration over depth:} Our primary aim is
        theorem--proof identification, prioritizing the combination of various modalities for a holistic approach over delving into advanced, singular modalities that may not offer comparative advantages. We start with base models, considering scalability and relevance to our scientific domain.

    \item \textbf{Budgetary limitations:} The computational cost of training and evaluating large models, especially across multiple modalities, necessitates a pragmatic approach. We test with base models due to their feasibility and the modular nature of our framework, which allows for flexibility in model choice and scaling.
\end{itemize}

\end{toappendix}


The intricacy of scientific terminology presents challenges for natural language processing models, necessitating domain-specific pretraining to improve their understanding of scientific language. Following insights from ~\cite{gururangan2020don}, which demonstrated performance gains through additional pretraining on diverse datasets ~\cite{acl-papers}, we adopt a tailored approach. Instead of extending an existing model like \textsf{RoBERTa}, we pretrain our model from scratch using a corpus of mathematical articles, aiming for a direct comparison with models trained on general English.
Note that final performance of the language model at our task is not our only
target: we are also interested in models that, after pretraining,
require fewer samples to fine-tune.

\paragraph*{Related work (text modality).}
Several existing works have built a domain-specific language model for
scientific papers, such as \textsf{SciBERT}~\cite{beltagy2019scibert},
\textsf{BioBERT} \cite{lee2020biobert}, and \textsf{MathBERT}~\cite{peng2021mathbert}.
\textsf{MathBERT} is pretrained on mathematical texts and formulas, showing notable efficacy in tasks like mathematical information retrieval and formula classification, albeit necessitating access to \LaTeX{} sources, unlike our PDF-based approach. \textsf{BioBERT} focuses on medical science, diverging from our focus, while \textsf{SciBERT} covers a broader spectrum, including computer science, making it a relevant baseline for our experiments. This comparison aims to assess the effectiveness of domain-specific pretraining in enhancing model performance with potentially less data, setting the stage for future exploration into more extensive, multi-billion parameter models.

Previous research~\cite{martin2019camembert}
has also underscored the significance of pretraining, showing that models
trained on just 4 GB of web-crawled data can outperform those trained on
over 130 GB, especially on domain-specific tasks.


\paragraph*{Methodology.}
We pretrain a language model from
scratch on a 50k vocabulary size (with byte-pair encoding), similar to the
configuration of \textsf{RoBERTa base (124M)}~\cite{liu2019roberta}.
While masking 15\% of tokens we kept the configuration similar to
original \textsf{RoBERTa} ($L=12, H=768, A=12$), but on a different
vocabulary. The model used dynamic masking and was trained on
masked language modeling loss.
After pretraining, the model is fine-tuned for our classification task.

\begin{toappendix}
We report pretraining results on two pretraining configurations (see
Table~\ref{table:Pretrain_config}): a \textsf{BERT}-like model in addition to the
\textsf{RoBERTa}-like model described in the main text.
As a quantitative measure of the quality of the pretraining, we report
the perplexity of the pretrained language model on the MLM task, similar
to Table 3 in the \textsf{RoBERTa} paper~\cite{liu2019roberta}. We show
the evolution of the MLM loss in
Figure~\ref{fig:mlm_loss}.
For a qualitative analysis, we intentionally picked up samples that require
specific vocabulary understanding on the MLM task, see Table~\ref{table:mlm_table}.
  \begin{table}[h]
      \small
\begin{tabular}{ cccccc }
 \midrule
 \bfseries Language model & \bfseries Batch size & \bfseries Steps
                          &\bfseries Learning rate& \bfseries Perplexity (epoch
 \#10)&\bfseries Time per epoch (h) \\
 \midrule
\bfseries \textsf{BERT}-like (110M) & 256 & 47\,773 & $2\times 10^{-5}$
 & 3.034 & 11 \\
\bfseries \textsf{RoBERTa}-like (124M) & 256 & 47\,773 & $2\times 10^{-5}$ & 2.857 & 16 \\
 \midrule
\end{tabular}
\caption{Pretraining configurations (on arXiv dataset)}
\label{table:Pretrain_config}
\end{table}

\begin{figure}
  \centering
  \includesvg[scale=.8]{images/mlm_pretrain_loss.svg}
    \caption{MLM loss for two pretrained models, as a function of the
    pretraining epoch}
    \label{fig:mlm_loss}
\end{figure}
\begin{table}
\begin{tabular}{p{.3\linewidth} p{.3\linewidth} p{.3\linewidth}}
 \toprule
 \textbf{Masked sentence}& \textbf{Pretrained \textsf{BERT}-like model}
                         &\textbf{\textsf{BERT} model}\\
 \midrule
 This concludes the [MASK].& proof
lemma
claim
theorem
case
thesis
&
game
story
film
play
episode
novel
\\
 \midrule
We show this by [MASK].
&
induction
.
definition
a
lemma
contradiction
&
ourselves
accident
name
themselves
ear
hand
\\
 \midrule
By [MASK]'s inequality.
&
jensen
holder
young
minkowski
cauchy
&
fourier
brown
russell
fisher
newton
\\
 \midrule
The [MASK] is definite positive.
&
{{inequality}}
case
slimit
sum
function
&
result
sign
value
answer
form
\\
 \midrule
In particular any field is a [MASK].
&
.
1
group
f
field
&
field
theory
domain
variety
category
\\
 \midrule
To determine the shortest distance in a graph, one can use [MASK]'s algorithm.
&
{{dijkstra}}
grover
tarjan
newton
hamilton
&
shannon
newton
taylor
wilson
moore
\\
 \midrule
An illustration of the superiority of quantum computer is provided by [MASK]'s algorithm.
&
{{grover}}
shor
dijkstra
yao
kitaev
&
turing
newton
shannon
maxwell
einstein
\\
 \midrule
One of the ways of avoiding [MASK] is using cross validation, that helps in estimating the error over test set, and in deciding what parameters work best for your model.
&
{{overfitting}}
error
errors
misspecification
noise
&
errors
error
this
bias
uncertainty
\\
 \midrule
\end{tabular}
\caption{Inference of \textsf{BERT}-like pretrained model on selected MLM tasks}
\label{table:mlm_table}
\end{table}

An example of use of Grad-CAM for visualization of the attention heads of a language model in Figure~\ref{fig:nlp_vis}.
\begin{figure}[htp]
\includegraphics{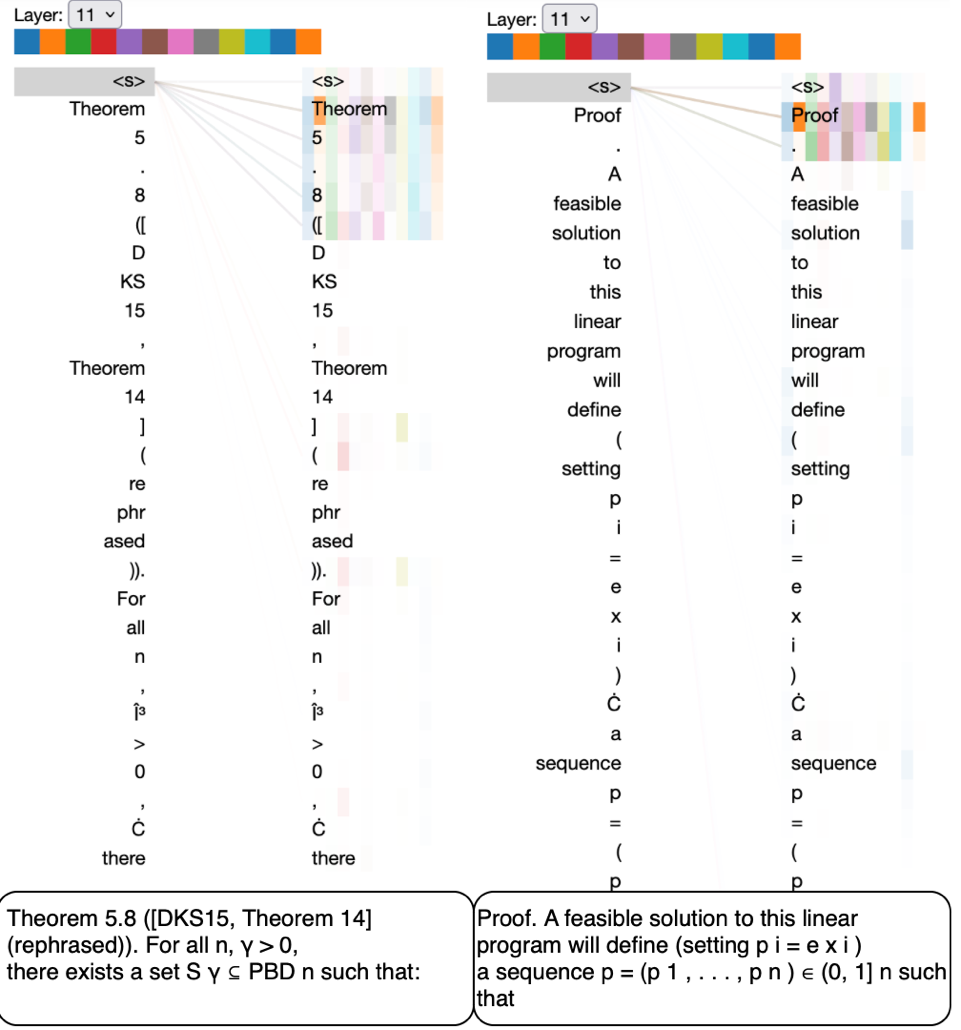}
\caption{Visualising the attention maps of a finetuned transformer Language model}
\label{fig:nlp_vis}
\end{figure}
\clearpage
\end{toappendix}

\subsection{Vision Modality}
\label{sec:vision}

\paragraph*{Related work (vision modality).}

Architecture design significantly influences model performance, evolving substantially from the early \textsf{Lenet-5}~\cite{lecun1998gradient}.

\textsf{ResNet} \cite{he2016deep} pioneered the use of skip connections, a concept expanded upon by \textsf{DenseNet} \cite{huang2017densely}, which connected each layer to all its predecessors. Following this, \textsf{NASNet} \cite{zoph2018learning} leveraged neural architecture search (NAS) for optimal architecture selection via reinforcement learning. Advancements continued with \textsf{EfficientNet} \cite{tan2019efficientnet}, which employed NAS to fine-tune hyperparameters and introduced compound scaling, significantly enhancing efficiency and performance over predecessors like \textsf{NASNet}, making it suitable for multimodal frameworks by minimizing computational demands.
The debut of vision transformers~\cite{dosovitskiy2020image} challenged
CNNs' supremacy, demonstrating superior results with ample data. However,
due to computational constraints, our focus remains on CNNs, particularly
after studies~\cite{tan2021efficientnetv2,liu2022convnet,woo2023convnext}
showed recent CNNs, including
\textsf{EfficientNetv2}~\cite{tan2021efficientnetv2}, achieving
comparable performance to transformers at a much lesser computational
cost.



Indeed, \textsf{EfficientNetv2} introduces a model family that
significantly outpaces its predecessor, \textsf{EfficientNet}, in
training speed and parameter efficiency on various datasets. It surpasses
the Vision Transformer (ViT)~\cite{dosovitskiy2020image} in performance
while maintaining a considerably smaller size. This efficiency makes
\textsf{EfficientNetv2M} — our chosen model — ideal for use as a network
backbone.
Key to its performance is the adoption of Fused-MBConv in early network stages, replacing depth-wise convolutions (MBConv) and optimizing for accuracy, parameter, and training efficiency through training-aware NAS. This approach, focusing on a $3\times 3$ kernel size while adding depth, ensures \textsf{EfficientNetv2}'s effectiveness even with the smaller image sizes of the ImageNet dataset, which typically challenges ViT models in training time and memory usage.

\paragraph*{Methodology.}


CNNs, pivotal in image classification and as backbones in visual-language tasks, typically benchmark on ImageNet and CIFAR for top-1\% accuracy.
Our project, targeting the identification of mathematical symbols and the
layout of paragraph blocks to discern proofs and theorems, necessitates
model training from scratch. Distinct markers like the term ``Proof'' in
unique fonts and the QED symbol, crucial yet overlooked by text
modalities, guide our focus.

\begin{toappendix}
\subsection{Vision Modality}
 Utilizing Grad-CAM~\cite{selvaraju2017grad}, we visually demonstrate the
 visual model's attention to specific elements such as ``Proof'',
 ``Theorem'' keywords r the use of italics, highlighting its effectiveness in recognizing mathematical documentation nuances (see Figure \ref{fig:grad_vis}).

 \begin{figure}[htp]
\includegraphics{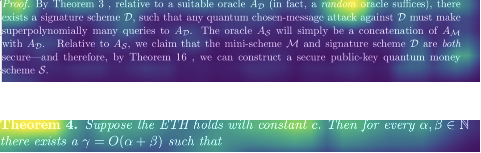}
\caption{Grad-CAM visualizations of some sample blocks}
\label{fig:grad_vis}
\end{figure}
\end{toappendix}

One specificity of vision approach for classification block is that
images come in widely different
\textbf{aspect ratios}. Traditional interpolation methods, though prevalent for adjusting natural images to a uniform resolution, unsuitably modify the geometry of text, symbols, and fonts in our context. Based on corpus analysis, we establish a fixed resolution of (400$\times$1400) pixels. This size accommodates over 80\% of our paragraphs, with larger images being cropped and smaller ones padded to maintain this standard without altering their intrinsic visual properties. This approach aligns with recommendations against scale variance \cite{touvron2019fixing} and parallels the preprocessing strategy used in the Nougat paper \cite{blecher2023nougat}, which also maintains a constant aspect ratio to suit specific model inputs. Our method ensures the preservation of textual image integrity by avoiding the pitfalls of resizing, opting instead for cropping or padding to fit our predetermined resolution criteria.

\begin{toappendix}
        In Figure~\ref{fig:cdf_hw}, we plot the cumulative distribution
        of heights and widths of paragraph blocks in our dataset, which
        is used to fix the common target resolution of all images.
    \begin{figure}[htp]
        \includegraphics[width=.475\linewidth]{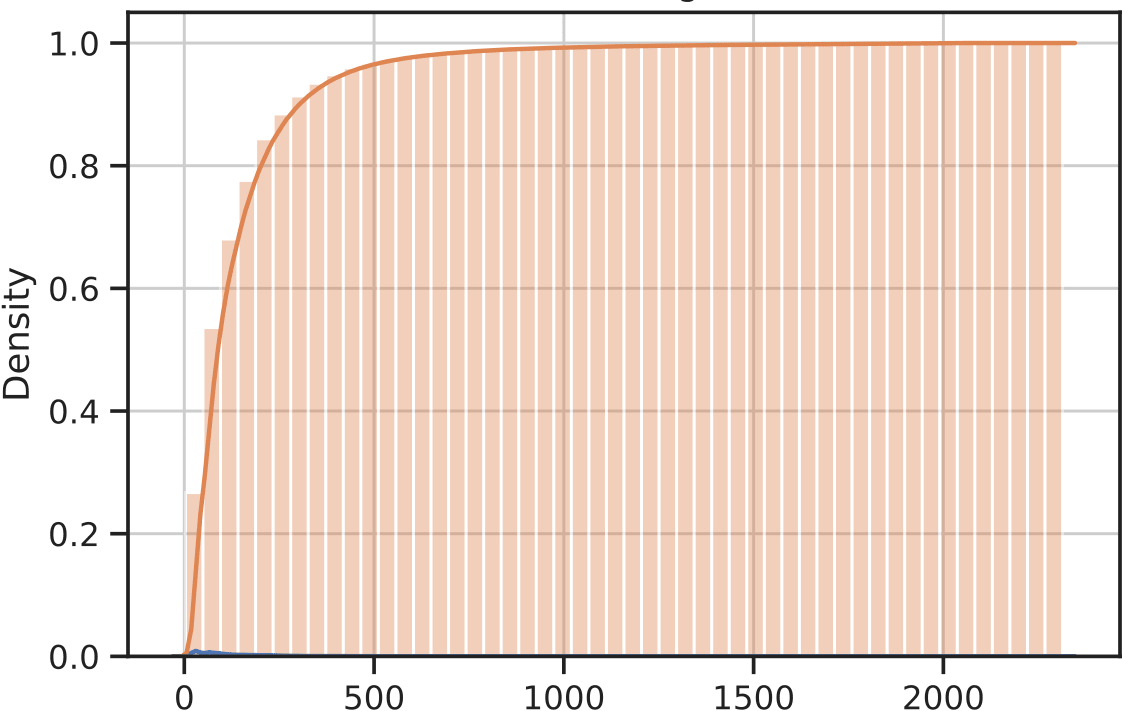}
        \hfill
        \includegraphics[width=.475\linewidth]{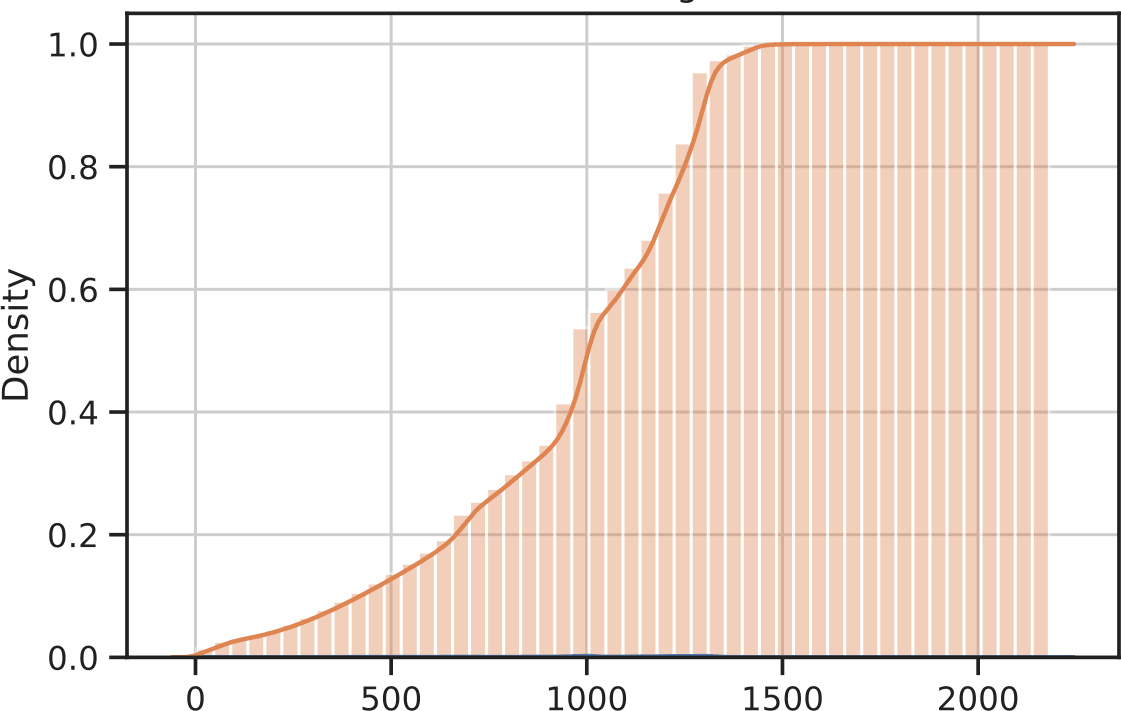}
        \caption{Cumulative distribution of heights (left) and widths (right) of paragraph blocks in
        our dataset}
        \label{fig:cdf_hw}
        \end{figure}

\end{toappendix}


To counteract the issue of \textbf{white
backgrounds} in scientific texts, which can hinder CNN performance as noted by studies \cite{hosseini2017limitation}, we invert image colors to mimic the MNIST dataset's white-on-black text presentation. This approach prevents max-pooling operations in CNNs from mistakenly prioritizing the background, thereby maintaining focus on the textual content.

\textsf{EfficientNet} comes with several variants (\textsf{B0}--\textsf{B7}) where B7 has the largest receptive field due to compound scaling. We select in our experiments a base network~(\textsf{B0}), a medium-sized network~(\textsf{B4}) and the largest network~(\textsf{B7}). \textsf{EfficientNetV2} also comes with different sizes. We focused on the small (\textsf{EfficientNetV2s}) and medium-sized (\textsf{EfficientNetV2m}) models.

\subsection{Font Modality}
\label{sec:font}

\begin{figure}
\hspace*{-.1\linewidth}\includegraphics[width=1.2\linewidth]{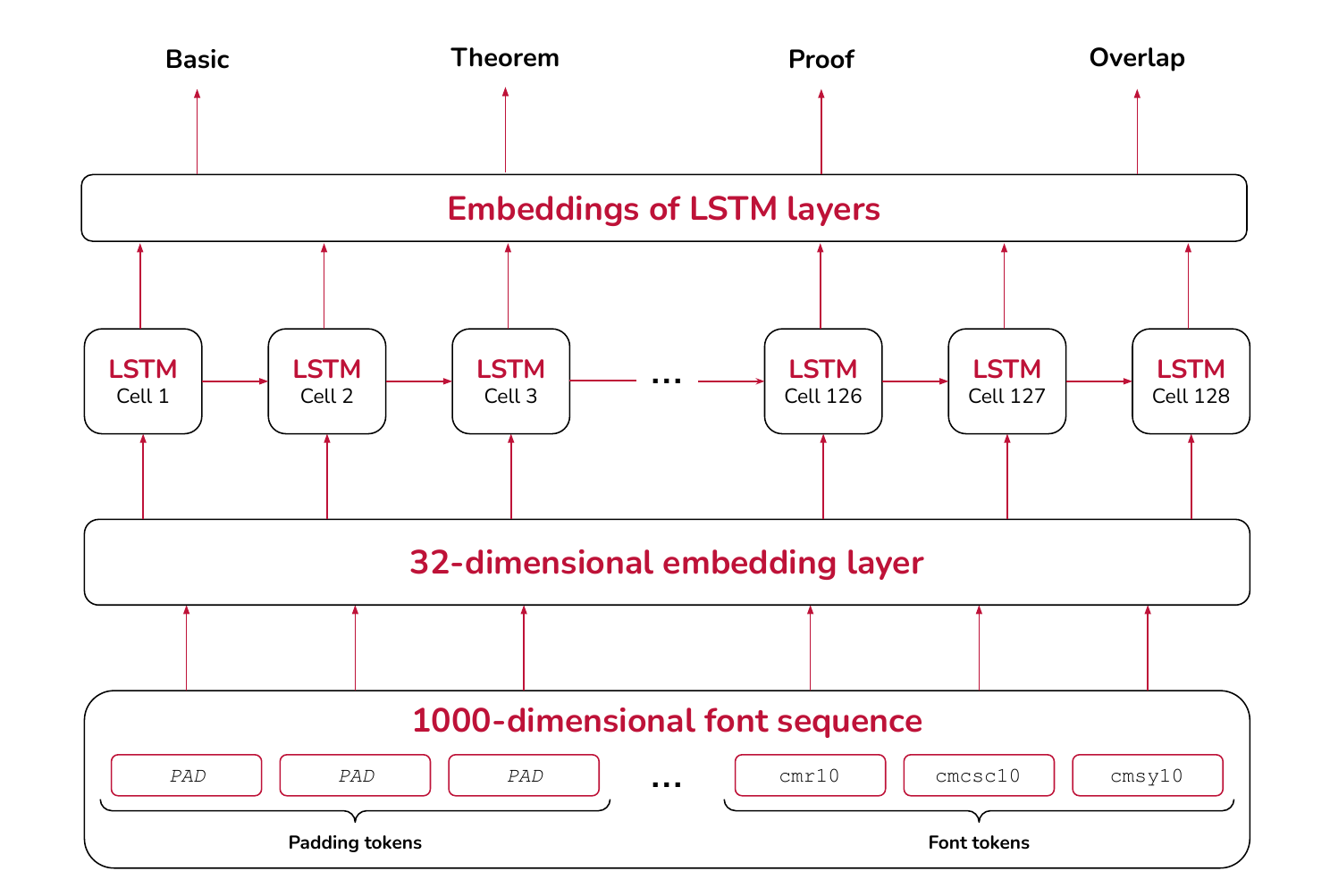}
\vspace{-2em}
\caption{LSTM model for the font modality}
\label{fig:lstm_font}
\end{figure}

The last modality we consider is styling information present in the PDF
in terms of the sequence of fonts (font family and font size) used in a
specific paragraph. This information can be obtained using the
\textsf{pdfalto} tool\footref{ft:pdfalto}, which produces a list of fonts used in a given
document, and associates each text token to a particular font. Fonts are usually standard \LaTeX{}
fonts, such as \verb|cmr10| for Computer Modern Roman in 10 point.

\begin{toappendix}
\subsection{Font Modality}
\begin{figure}
\includesvg[width=\linewidth]{images/font_inalto.svg}
\caption{Font information as presented in the output of the
  \textsf{pdfalto} tool. Note that, for instance, the $\mathbb{Z}$ or $m$
characters of the title are written in different fonts from the other
tokens.}
\label{fig:font_prep}
\end{figure}
We illustrate in Figure~\ref{fig:font_prep} the output of
\textsf{pdfalto} on an example PDF document, highlighting the font
sequence information.
\end{toappendix}

From the training data, we build a font vocabulary of $4\,031$ unique
fonts including their sizes, and represent every paragraph block as a
sequence of font identifiers.
To match input dimensions among training samples, we apply left padding
with a maximum length of 1\,000. We then feed the entire sequence to a
simple 128-cell LSTM~\cite{hochreiter1997long}  network to monitor
the loss, represented in Figure~\ref{fig:lstm_font}.
The choice of the model is purely
to capture sequential information within fonts that can be used to
identify the label of the paragraphs.

\section{Multimodal and Sequential Models}
\label{sec:beyond-unimodal}
We now go beyond unimodal models by showing how all three modalities can
be combined into a single late-fusion multimodal model, and how block
sequence information can be captured.

\paragraph*{Related work (Multimodality).}

Multimodal machine learning for document AI has seen a surge in interest.
However, existing models often fall short in addressing the unique
aspects of scientific articles, such as font features, scientific
terminology, and the structure of lengthy documents. Most research
focuses on benchmarks like FUNSD~\cite{jaume2019},
CORD~\cite{park2019cord}, and RVL-CDIP~\cite{harley2015icdar}, which deal
with simpler document types like invoices and forms assuming dependencies
within the same page.




Several architectures have been proposed as multimodal transformers which
try to jointly model different modalities in a single transformer model
early on during the input stage and try to capture modality interactions
such as LayoutLM~\cite{xu2020layoutlm,xu2020layoutlmv2,huang2022layoutlmv3}, which takes into account 2D positional embedding via masked visual-language model loss and multi-label document classification loss.
An alternative approach
is late fusion (after feature extraction) such as CLIP
\cite{radford2021learning}. In CLIP, both text and visual features are
projected to a latent space with identical dimensions and a contrastive
loss is applied to  zero shot learning with supervision from
language models. One of the big advantages of CLIP is its ability to
upgrade and replace the backbone on the fly.

We adopt a late fusion based approach (similar to CLIP) instead of early
fusion based approaches such as LayoutLM
for the
following reasons:
\begin{inparaenum}[(i)]
    \item \textbf{Modular backbone integration:} Our late fusion approach
        is driven by the flexibility to integrate various backbones in a
        modular way, enhancing performance and scalability without being constrained by fixed architecture dimensionality.
    \item \textbf{Reevaluating cross-modality capture:} the specialized
        losses for cross-modal interactions, like those in
        \textsf{LayoutLMv2} and \textsf{LayoutLMv3}, are claimed to
        enhance cross-modality relationship understanding. However, this
        assumption warrants further scrutiny. Specifically,
        the specificities of \textsf{LayoutLM} architectures mean that they cannot
        be compared to straightforward multimodal fusion strategies employing identical backbones.
        Here, we conduct direct comparisons across multiple fusion methods, focusing on raw features and employing only cross-entropy classification loss, while maintaining consistent backbones as in unimodal setups.
\end{inparaenum}

\begin{toappendix}
\paragraph*{Comparison to SOTA models (in Doc AI)}
LayoutLM's different versions use coordinate information at the token
level after OCR (see Figure 2 of the LayoutLM paper), enabling
token-level classification or single-label assignment to a page's
content. This approach can’t natively label specific paragraphs based on
logical structure from Grobid. For tasks available with LayoutLM, see
available model heads in LayoutLM's HuggingFace API documentation.

Donut and Nougat employ an encoder--decoder architecture (mainly for
generative or Q\&A tasks), while our approach is encoder-only (assigning
labels to fixed paragraphs). This makes direct comparison challenging due
to the generative nature of decoder models.
We compare our models with Hierarchical Attention Transformer (HAT), specifically suited for long documents, as they are encoder-only and can label paragraphs effectively. We also provide HATs with multimodal capability.
            \paragraph*{Details on LayoutLM comparisons to baselines.}
            In the LayoutLM study, Table 5
            showcases comparisons with various baselines for document
            classification tasks, many of which are unimodal, alongside a
            single multimodal approach adapted from
            \cite{dauphinee2019modular}. Successive iterations,
            \textsf{LayoutLMv2} and \textsf{LayoutLMv3}, primarily
            benchmark against the initial version or this same multimodal
            baseline, as evidenced in Table 3 of \cite{xu2020layoutlmv2}
            and Table 1 of \cite{huang2022layoutlmv3}. This pattern of
            comparison is echoed in other document classification
            frameworks like \textsf{LiLT} \cite{wang2022lilt}, where the
            multimodal baseline incorporates significantly less powerful
            backbones. The baseline for multimodal comparison utilizes an
            XGBoost classifier \cite{chen2016xgboost}, processing class
            scores (not features) from basic backbones (\textsf{VGG-16}
            \cite{simonyan2014very} for visual and \textsf{BOW} for text), a setup
            that superficially engages with multimodality compared to
            \textsf{LayoutLM}'s use of \textsf{ResNet-101} and
            \textsf{BERT}. Despite employing less
            advanced backbones, the performance of this multimodal
            network (93.03, as seen in Table 5 of the \textsf{LayoutLM} paper)
            closely approaches that of \textsf{LayoutLM} (94.42). This observation
            raises critical questions about the source of
            \textsf{LayoutLM}'s performance gains: Are they due to its unique loss functions, or merely the result of employing stronger baseline architectures for comparison?

            \paragraph*{Related Work on \textsf{DiT}.}
            \textsf{DiT} \cite{li2022dit}, a notable architecture for
            document classification leveraging a vanilla Vision
            Transformer (\textsf{ViT}) backbone, is showcased for its
            adaptability to classification tasks in Table 1 of its
            publication. This table, however, limits its comparison to
            \textsf{DiT}'s performance (92.11) against the
            \textsf{ResNext} model (90.65), which we consider a
            suboptimal baseline due to advancements in CNN models like
            \textsf{EfficientNet} and \textsf{EfficientNetv2}. These
            newer models not only enhance performance but also optimize
            training and inference times. For context, Table~2 in the
            \textsf{EfficientNet} paper \cite{tan2019efficientnet} and Table 7 in the
            \textsf{EfficientNetv2} paper \cite{tan2021efficientnetv2}
            offer direct comparisons with the \textsf{ResNext} and
            vanilla \textsf{ViT} backbones, respectively, demonstrating
            the superiority of \textsf{EfficientNet} variants. Notably,
            Table 1 of the \textsf{DiT} paper \cite{li2022dit} omits
            metrics such as FLOPs or training time per epoch, details
            that are explicitly addressed in Table 7 of
            \cite{tan2021efficientnetv2}, underscoring \textsf{EfficientNetv2}'s advancements over \textsf{ViT}-based architectures.
        \end{toappendix}



\paragraph*{Methodology.}

We compare different modalities for late fusion: bilinear gated
units~\cite{kiela2018efficient},
\textsf{EmbraceNet}~\cite{choi2019embracenet}, gated multimodal units
(GMU) \cite{arevalo2017gated}, and attention mechanisms such as those of \textsf{ViLBERT} \cite{lu2019vilbert}, typically applied to dual modalities but extendable to multiple. These methods, focusing on feature-level fusion, ensure modularity and adaptability across architectures.

Our comparison to simple fusion methods is narrowed to concatenation, identified as the most effective among basic fusion strategies, allowing direct comparison with our unimodal baselines. These comparisons solely rely on cross-entropy loss for classification tasks, omitting additional losses like contrastive loss. Importantly, the feature backbones are frozen during fusion, preventing weight updates and situating multimodal fusion as an augmentation to our unimodal framework.

\begin{figure}
\includegraphics[width=\linewidth]{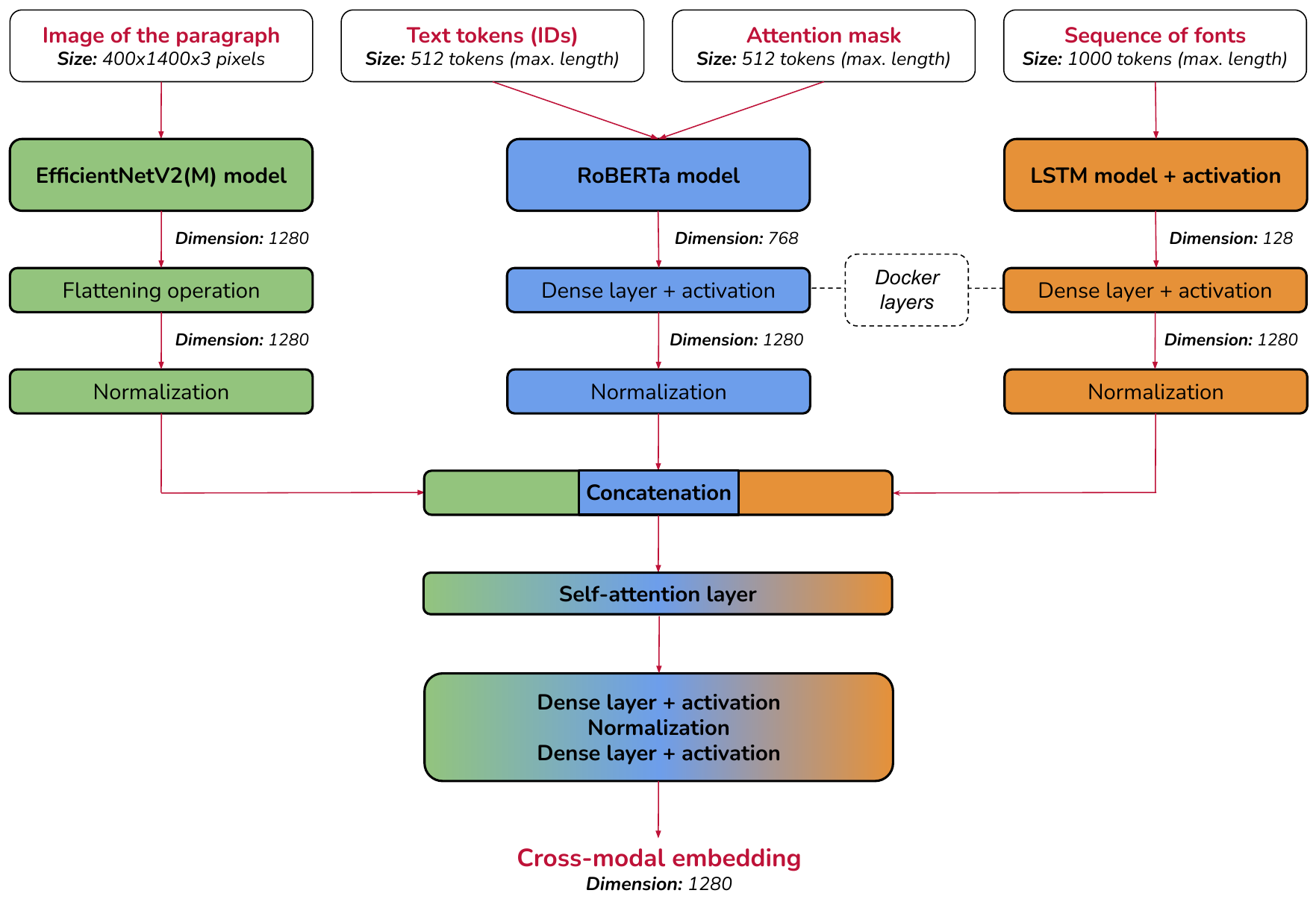}
\caption{Cross-modal attention architecture}
\label{fig:cross_modal}
\end{figure}

    \paragraph*{Cross-modal attention architecture.} The main multimodal
    model we use is a cross-modal attention model, inspired by
    \textsf{ViLBERT}'s attention mechanism. We show its full architecture
    in Figure~\ref{fig:cross_modal}.

\paragraph*{Sequential approach.}
\label{sec:crf}

In addition to modalities, considering the sequencing of the blocks,
i.e., the order in which they appear in the document, allows us to
determine with greater confidence the class of each block. For example,
if one seeks to predict the class of a block that is itself framed by two
blocks that have been classified as \emph{proof}, then there is a good chance
that this block is itself a \emph{proof}. We consider how to integrate
unimodal and multimodal models into a sequential prediction model.

To classify scientific paragraphs spanning multiple pages, long
document classification methods like Tobert, Bigbird, Longformer,
and Hierarchical Attention Transformer (HAT) are more relevant.
HAT outperforms Bigbird and Longformer (see Table 5 of HAT pa-
per) and efficiently handles long documents, including page breaks.
However, these models lack multimodal capabilities and document-
specific information, such as coordinate data used in LayoutLM
and LILT. Our approach combines the strengths of both classes
while using the same backbones. It is computationally efficient,
relying on the SW mechanism instead of full attention, and saves
nearly half the parameters per encoder by reducing the feedforward
dimension.

\begin{figure*}
\hspace*{-.05\linewidth} \includegraphics[width=\linewidth]{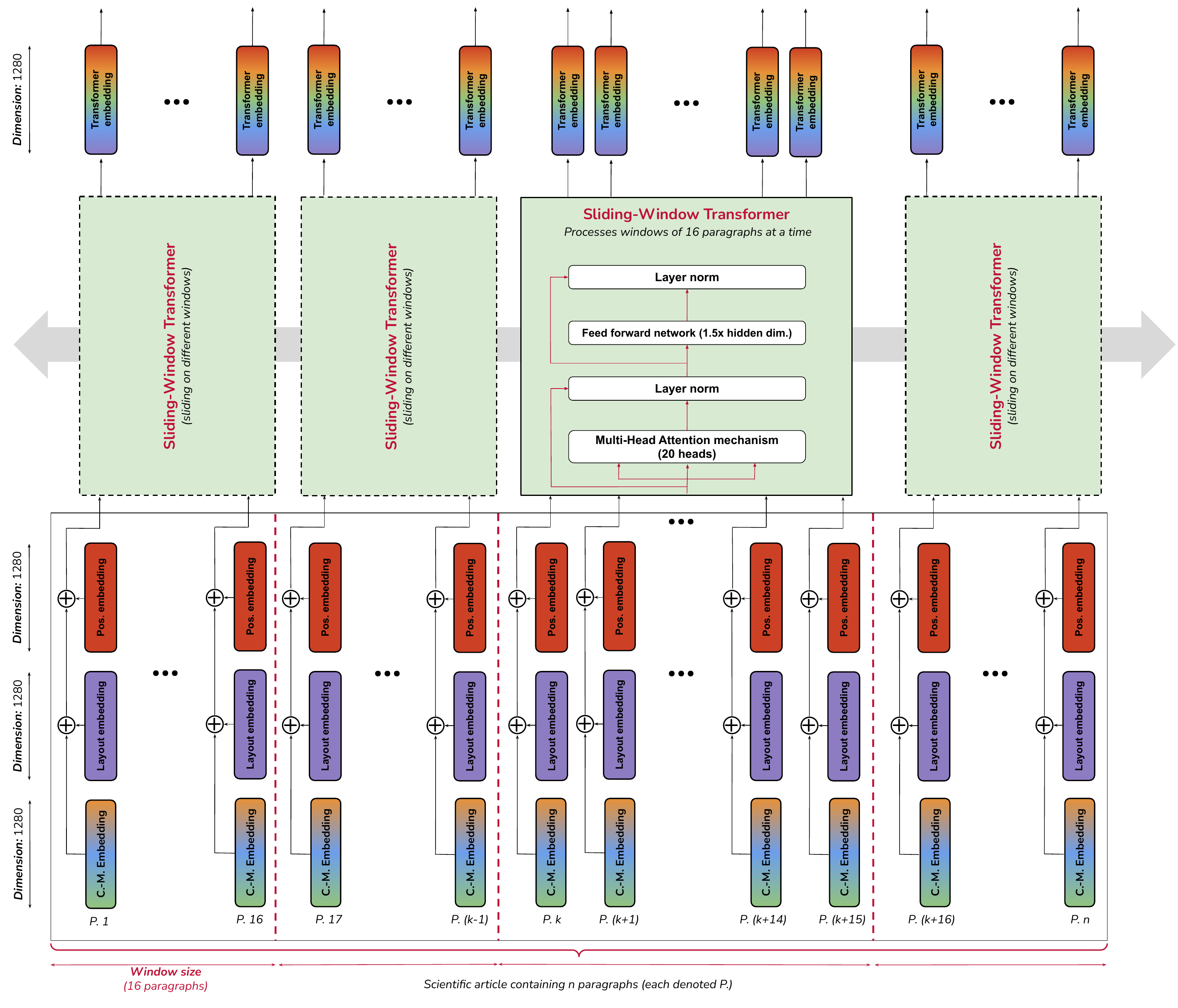}
\vspace{-2em}
\caption{Sequential model based on a sliding-window (SW) transformer architecture}
\label{fig:sliding_window}
\end{figure*}

We propose two approaches to do this:
First, using a simple
linear-chain order-one Conditional Random Fields model
(CRFs)~\cite{CRF2001Lafferty}.
Second, we introduce a novel transformer-based BERT-like encoder architecture (also more efficient for our task) to
process multimodal features, using a sliding window (SW) of size~$k=16$, whose
architecture is presented in Figure~\ref{fig:sliding_window}. We also investigate the impact of long sequential relationships by employing interleaving architecture found in Hierarchical Attention Transformers (HATs) \cite{chalkidis2022exploration}. The architecture is modified to be adapted in a multimodal setting such as ours.

The CRF and SW models use the following features, on top of frozen unimodal or multimodal
model:
unimodal text, vision, and font models respectively bring 768, 1280,
and 128 features;
the multimodal approach includes 1280 joint features;
we incorporate four additional geometrical features to describe block positions: normalized page number, indicating a block's page relative to the total pages; normalized horizontal and vertical distances from the block's bounding box corners; and a binary feature indicating if a block and its predecessor are on the same page.

\begin{toappendix}
    \paragraph*{Details on transformer architecture for sequential
    approach.}
Our investigation leverages transformer-based BERT like architecture to
process multimodal features (1280 +6 layout features), enhancing the
traditional model to support the complex structure of PDF documents in
our dataset. By adjusting the ``maxlen'' parameter, our approach accommodates very large document lengths (see data distribution, Figure \ref{fig:Dataset_distrib}), significantly exceeding standard transformer configurations. This modification involves splitting longer documents and integrating layout embeddings, including page information, to enhance the attention mechanism across paragraphs, thus addressing the challenge of extensive document lengths.

\begin{figure}[htp]
\centering\includesvg[width=.5\linewidth]{images/disb_train_val.svg}
\caption{Overview of dataset distribution (number of paragraphs in each
PDF in the training and validation data)}
\label{fig:Dataset_distrib}
\end{figure}

Empirical evidence suggests that dividing PDFs into smaller segments or
applying a focused attention mechanism on a small window substantially
improves model generalization (see Tables \ref{tab:max_len_performance}
and \ref{tab:architecture_optimization}). Contrary to initial
expectations, increasing the encoder's complexity did not proportionally
enhance performance (see Table \ref{tab:encoder_block_performance}). A
strategic reduction in the feedforward network's dimensionality, from the
suggested fourfold in BERT paper to 1.5 times, not only elevated accuracy
but also streamlined the model's architecture, (see
Table~\ref{tab:transformer_comparison}), demonstrating efficiency gains alongside performance improvements.

In response to the discerned improvement in transformer models when
handling shorter data segments, our research pivots towards employing a
sliding window mechanism (as described in Figure
\ref{fig:sliding_window}). This method processes sequences of uniform
size, applying padding as required, and functions via a non-overlapping
approach, thereby enhancing computational efficiency by lowering the
complexity of token attention from $O(N^2)$ to $O(N \times k)$ where k
denotes the window size. This shift not only facilitates data
preprocessing by removing the need for manual data segmentation but also
introduces window size as a pivotal hyperparameter, substantially
improving performance across diverse ``maxlen'' configurations. The
implementation of this sliding window technique necessitated
modifications in training duration; specifically, training epochs were
doubled for each incremental doubling of ``maxlen'', maintaining consistent performance across varying data lengths. This strategic adaptation ensures model robustness and accuracy, even with extended document paragraphs (see Table \ref{tab:lower_maxlen_impact}).
We also make comparisons with Hierarchical Attention Transformers but do
not find any significant performance gains (see Tables \ref{tab:model_performance_reps3} and \ref{tab:model_performance_swe_cwe}) over a simple sliding-window transformer.
\end{toappendix}

In order to determine whether long-distance dependencies are also useful
to capture for our task, we also implement HATs, relying on the same
Sliding Window transformer encoder architecture used as as a
segment-wise encoder. We then expanded it to learn about connections
between different context windows (using cross segment encoder) taking
only the Multimodal [CLS] token of every segment. Out of the many
versions proposed in the original HAT paper
\cite{chalkidis2022exploration}, we tested the best-performing one,
i.e., with interleaving layers. See Figure~\ref{fig:hat_arch} for the
corresponding architecture.

\begin{figure}[htp]
\includesvg[width=\linewidth]{images/hat_layers_new.svg}
\caption{HAT network visualization (with 1 interleaving layer)}
\label{fig:hat_arch}
\end{figure}

\begin{toappendix}
    \paragraph*{Hierarchical Attention Transformers.}
    The intricate process of sequentially labeling paragraphs benefits
    from the nuanced understanding of adjacent textual relationships, a
    task adeptly managed by the sliding-window transformer model.
    Nevertheless, this approach may not fully account for long-term
    dependencies and interactions across windows, elements crucial for
    comprehensive document classification tasks where understanding
    global document context is paramount. To navigate these complexities,
    models like
    \textsf{RoBERT}/\textsf{ToBERT}~\cite{pappagari2019hierarchical},
    \textsf{Longformer}~\cite{beltagy2020longformer}, and
    \textsf{BigBird}~\cite{zaheer2020big} have been developed,
    specifically designed to bridge this gap.
    \textsf{RoBERT}/\textsf{ToBERT} enhances the sliding window framework
    with additional layers to capture wider textual relationships, while
    \textsf{Longformer} and \textsf{BigBird} refine the attention mechanism to balance local and global textual insights effectively.

    Building on these advancements, the Hierarchical Attention
    Transformer Network (\textsf{HAT})~\cite{chalkidis2022exploration}
    employs a layered approach, utilizing transformer encoders to forge a
    deeper connection between separated windows. Diverging from
    \textsf{Longformer} and \textsf{BigBird}'s emphasis on modified
    attention mechanisms, \textsf{HAT} leverages a series of encoder blocks to methodically process multimodal features across stacked sliding windows, thus addressing long-term dependencies more comprehensively.

    In our exploration, \textsf{HAT} model integrates two distinct types
    of encoder blocks: the Sliding Window Encoder (SWE) for encoding within-window modal information and the Context-Wise Encoder (CWE) for bridging content across windows. While SWE hones in on local attention, CWE extends its reach to encompass a broader context, employing an architecture designed to facilitate cross-window communication.

    \begin{table}[ht]
    \centering
    \caption{HAT Performance with \textbf{2 interleaving layers} (SWE=1,
    CWE=1, and  for Max Len 1024 and 32 epochs, similar to the training
configurations reported in Table \ref{tab:lower_maxlen_impact})}
    \label{tab:model_performance_swe_cwe}
    \begin{tabular}{lccccc}
    \toprule
    \textbf{Window Size} & \textbf{Params (M)} & \textbf{Train Loss}   & \textbf{Accuracy (\%)} &  \bfseries Mean $\textrm{F}_1$ (\%) \\
    \midrule
    16 & 47 & 0.3465 & 86.20 & 85.80 \\
    32 & 47 & 0.3415 & 86.58 & 85.93 \\
    64 & 47 & 0.3006 & 86.44 & 85.47 \\
    128 & 47 & 0.2990 & 85.18 & 84.10 \\
    256 & 47 & 0.3279 & 87.52 & 86.58 \\
    512 & 47 & 0.5040 & 79.81 & 78.03 \\
    \bottomrule
    \end{tabular}
    \end{table}

    \begin{table}[ht]
    \centering
    \caption{HAT Performance with \textbf{3 interleaving layers} (SWE=1,
    CWE=1, and  for Max Len 1024 and 32 epochs, similar to the training
configurations reported in Table \ref{tab:lower_maxlen_impact})}
    \label{tab:model_performance_reps3}
    \begin{tabular}{lcccc}
    \toprule
    \textbf{Window Size} & \textbf{Params(M)} & \textbf{Train Loss} &  \textbf{Accuracy (\%)} &  \bfseries Mean $\textrm{F}_1$ (\%) \\
    \midrule
    16 & 74 & 0.2996 & 86.01 & 85.15 \\
    32 & 74 & 0.2917 & 86.43 & 85.78 \\
    64 & 74 & 0.2758 & 86.70 & 86.05 \\
    128 & 74 & 0.2871 & 86.62 & 85.90 \\
    256 & 74 & 0.4460 & 80.45 & 78.96 \\
    512 & 74 & 0.5304 & 76.95 & 75.46 \\
    \bottomrule
    \end{tabular}
    \end{table}

    Despite the potential of \textsf{HAT} in enhancing model performance
    through structural and attentional depth, our investigations reveal
    that, within our multimodal framework, the added complexity of
    \textsf{HAT} does not unequivocally translate to superior performance
    over simpler sliding window constructs, highlighting a nuanced
    balance between architectural innovation and task-specific efficacy.
    We implement the interleaving architecture (as denoted in Figure 3b
    of \textsf{HAT} paper \cite{chalkidis2022exploration}.)

    To investigate the interleaving \textsf{HAT} we operate on variable
    hierarchical window sizes similar to the parameter sliding window
    (see Tables \ref{tab:model_performance_swe_cwe} and \ref{tab:model_performance_reps3})

\end{toappendix}

\section{Dataset and Setup}\label{sec:dataset}

We use
\textsf{Grobid}\footnote{\url{https://github.com/kermitt2/grobid}}
\cite{grobid}, which is the state of the art
for information extraction from scholarly documents to
parse a PDF document and interpret it
into a succession of paragraph blocks.

\begin{figure}
\hspace*{-.07\linewidth}\includegraphics[width=1.14\linewidth]{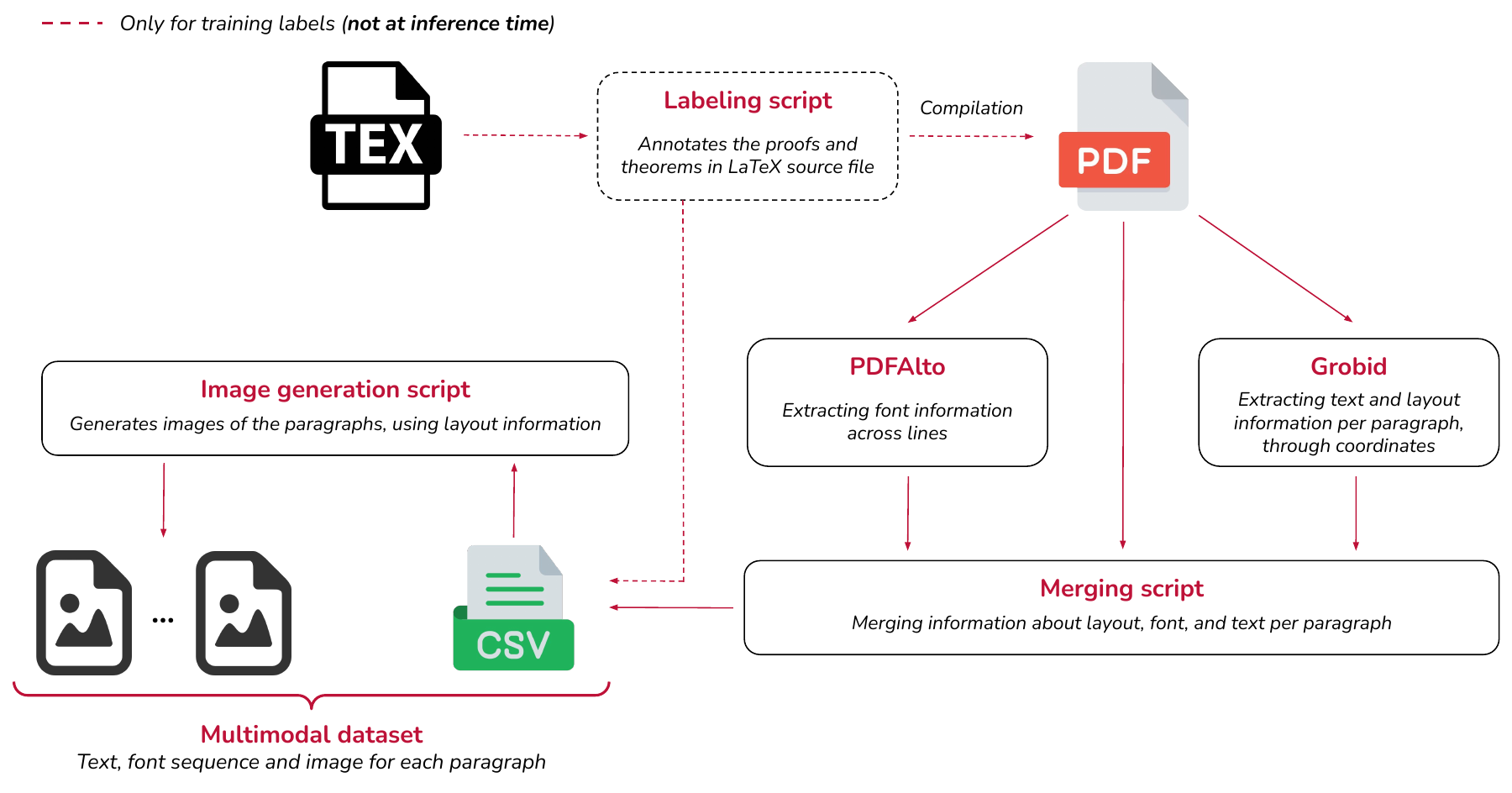}
  \vspace{-2em}
\caption{Dataset preparation pipeline}
\label{fig:Dataset_prep}
\end{figure}

Our dataset, encompassing all arXiv papers (around 1.7 million papers) up
to May 2020, was acquired via arXiv's bulk data access on Amazon S3. We
developed an annotation script to pinpoint theorem-like environments and
proofs within these documents, leveraging \LaTeX{} sources. This involved
crafting a \LaTeX{} package to instrument commands such as \verb|\newtheorem|
for precise identification in the compiled PDFs ($\approx$ 460k papers).
See Figure~\ref{fig:Dataset_prep}.
We filtered articles from the dataset to only keep those in English, for which
\LaTeX{} source is available (according to arXiv's policy, all
those that have been produced using \LaTeX{}), that were compilable on a
modern \LaTeX{} distribution, that contained at least a
theorem or a proof environment, and for which none of the tools
(our ground-truth annotation package, \textsf{Grobid} for extraction of
blocks, \textsf{pdfalto} for line-by-line font sequences, bitmap image
rendering for CNN's) failed to produce a valid output. This resulted in a
final dataset of $\approx$~197k papers. We stress that
\LaTeX{} sources are only used to produce ground-truth annotations, they are
not required at inference time.
\textsf{Grobid} sometimes fails to
extract correct paragraphs, i.e., some of the paragraphs
identified by \textsf{Grobid} overlap blocks of different category (say,
\emph{basic} and \emph{theorem}). We label such paragraphs as
\emph{overlap}, exclusively used for such outliers.

Our validation set comprises approximately 500\,000 paragraph blocks from
3\,682 randomly selected PDF articles. The remaining articles formed the training dataset, used entirely for pretraining our language model after filtering potential personal information such as author names and institutions from Grobid extractions to minimize privacy concerns.
Training involved dividing the dataset into batches of 1\,000 PDF
articles, incrementally fitting classifiers on these batches until
convergence, without exceeding a few dozen batches. Post-training,
classifiers' weights were frozen for integration into the multimodal
classifier, subsequently employed as feature extractors for the
sequential approaches detailed in Section~\ref{sec:crf}.
The dataset is heavily imbalanced, with
the number of paragraphs labeled as \emph{basic}: 314\,501, \emph{proof}:
125\,524, \emph{theorem}: 85\,801, and \emph{overlap}: 3\,470.

All experiments were run on a supercomputer with access at any point to 4
NVIDIA (V100 or A100) GPUs. We estimate to 8\,000 GPU hours the
computational cost of the entire prototyping, hyperparameter tuning,
training, validation, and evaluation pipeline.

\section{Experimental Results}
\label{sec:experiments}

We now report experimental results on the
\emph{basic}--\emph{theorem}--\emph{proof} classification problem, first
comparing representative unimodal classifiers, with and without the article paragraphs fed to the sequential approach, followed by the multimodal classifier. We then delve into more specific details of every unimodal classifier.

\begin{table*}[tb]
\centering
\caption{Overall performance comparison (accuracy and mean $\textrm{F}_1$ over the
  three classes \emph{basic}, \emph{theorem}, and \emph{proof})
  of individual modality models and
multimodal model, with and without the sequential approach; for
each model, the number of batches (1\,000 PDF documents, roughly 200k
samples) it was trained on is indicated (here + indicates additional batches on which further training of sequential paragraph model)}
\vspace{-1em}
\hspace*{-2em}\begin{tabular}{>{\bfseries}l l c c c c c}
  \toprule
  \bfseries Modality & \bfseries Model chosen & \bfseries Seq.~approach & \bfseries \#Batches & \bfseries \#Params (M) & \bfseries Accuracy (\%) & \bfseries Mean $\textrm{F}_1$ (\%) \\
  \midrule
  Dummy & always predicts \emph{basic}& --- & --- & --- & 59.41 & 24.85 \\
    Top-$k$ first word & use only first word & --- & --- & --- & 52.84 & 44.20 \\
    Line-based \cite{mishra2021towards} & Bert (fine-tuned) & --- & --- & 110 & 57.31 & 55.71 \\
  \midrule
  \multirow{3}{*}{Font} & \multirow{3}{*}{LSTM 128 cells} & - & 11\phantom{+0} & 2 & 64.93 & 45.48  \\
                        & & CRF & 11+8 & 2 & 71.50 & 64.51 \\
                        & & SW Transformer & 11+8 & 2 & 76.22 & 71.77 \\
  \midrule
  \multirow{3}{*}{Vision} & \multirow{3}{*}{EfficientNetV2m\_avg} & - & 9\phantom{+0} & 53 & 69.44 & 60.33 \\
                          & & CRF & 9+8 & 53 & 74.63 & 70.82 \\
                          & &SW Transformer & 9+8 & 65 & 79.59 & 77.66 \\
  \midrule
  \multirow{3}{*}{Text} & \multirow{3}{*}{Pretrained \textsf{RoBERTa}-like} & - & 20\phantom{+0} & 124 & 76.45 & 72.33 \\
                        & & CRF & 20+8 & 124 & 83.10 & 80.99 \\
                        & &SW Transformer & 20+8 & 129 & 87.50 & 86.67 \\
  \midrule
  \multirow{3}{*}{Multimodal} & \multirow{3}{*}{Cross-modal attention} & - & 2\phantom{+0} & 185 & 78.50 & 75.37 \\
                              & & CRF & 2+8 & 185 & 84.39 & 82.91 \\
                              & &SW Transformer & 2+8 & 198 & \bfseries 87.81 & \bfseries 87.18 \\
                               & &HAT 
                               & 2+8 & 232 &  87.52 &  86.58 \\
  \bottomrule
\end{tabular}%
\label{tab:crf_results}
\end{table*}

\begin{table*}
  \centering\small
    \caption{Performance comparison of multimodal fusion techniques (with @dimensions and model architecture)}
\vspace{-1em}
  \hspace*{-2em} 
  \begin{tabular}{lrcrcc}
    \toprule
    \textbf{Model Architecture} & \textbf{\#Params (Total/Trainable)} & \textbf{Accuracy (\%)} & \textbf{Mean $\textrm{F}_1$ (\%)} \\
    \midrule
    Concatenated raw features(@2176) & 179M/8K & 77.90 & 74.34 \\
    docker layers(@1280) + concat(@3840) & 180M/1M & 78.11 & 74.95 \\
    docker layers(@1280) +fusion (@768) & 183M/4M & 78.50 & 75.38 \\
    docker layers(@1280) +fusion (@1280) & 185M/6M & 78.43 & 75.24 \\
    docker layers(@1280) +fusion (@2304) & 189M/10M & 78.42 & 75.13 \\
    bilinear mechanism (@1280) & 182M/3M & 77.99 & 74.78 \\
    docker layers (@1280) +bilinear gated mechanism(@1280) & 185M/6M & 78.30 & 75.11 \\
    docker layers(@1280) +GMU mechanism (@1280) & 185M/6M & 78.11 & 75.52 \\
    \textbf{docker layers (@1280) +attention mechanism (@1280)} & \textbf{185M/6M} & \textbf{78.50} & \textbf{75.37} \\
    docker layers(@1280) +multihead attention (@1280, 8 heads) & 244M/65M & 78.33 & 75.26 \\
    EmbraceNet mechanism (@1280) (balanced prob +docker layers (@1280) incl) & 182M/3M & 77.73 & 74.70 \\
    EmbraceNet mechanism (@1280)(weighted prob +docker layers (@1280) incl) & 182M/3M & 77.73 & 74.55 \\
    docker layers(@1280) +fusion (@2304) +fusion (@768) & 191M/12M & 78.50 & 75.32 \\
    docker layer (@1280) +Cross-modal attention (@1280) +fusion (@768) & 186M/7M & 78.45 & 75.24 \\
    docker layer (@1280) +GMU mechanism (@1280)+fusion (@768) & 186M/7M & 78.33 & 75.28 \\
    \bottomrule
  \end{tabular}
  \label{tab:model_comparison}
\end{table*}

We are interested in two main performance metrics: \emph{accuracy}
measures the raw accuracy of the classifier on the validation dataset
(disjoint with the training dataset); and (unweighted arithmetic)
\emph{mean $\textrm{F}_1$-measure} of the \emph{basic}, \emph{theorem}, and
\emph{proof} classes, which summarizes the precision and recall over each
class assigning the same weight to every class. As \emph{basic} is the
most common class in the dataset, a \emph{dummy} classifier that would
always predict the \emph{basic} class would have an accuracy of
$59.41\%$; but its recall would be $100\%$ on \emph{basic} and $0\%$ on
the other classes, while its precision would be $59.41\%$ on \emph{basic}
and $0\%$ on the other classes, resulting in a mean $\textrm{F}_1$ of
\(\frac{1}{3}\times \frac{2\times 59.41\%}{59.41\%+100\%}\approx
24.85\%\). This gives an important comparison point for all other
methods; accuracy measures how well the classifier works on the actual
unbalanced data, while mean~$\textrm{F}_1$ favors methods performing well to
identify all three classes, arguably a better metric.

Drawing inspiration from two related works, albeit applied in slightly different settings, we evaluate two straightforward baselines:
\begin{inparaenum}
    \item Top-$k$ first words: This method, which echoes the approach used
      in \cite{DBLP:journals/corr/abs-1908-10993} focusing on the first
      paragraph of marked environments, constructs a vocabulary of the
      top-$k$ unique words for each class. Labels are assigned based on
      the first word of a text and whether it matches any word within the
      class-specific vocabulary. For instance, if the first word is
      within \{{\textit{theorem, lemma, proposition,
      definition}}\}, the text is labeled as a theorem.
    \item Text classifier from~\cite{mishra2021towards}: We reuse the
      text classifier that was fine-tuned in~\cite{mishra2021towards},
      which processes text lines (not paragraphs) extracted from pdfalto. Note this
      classifier does not identify the \emph{overlap} class.
\end{inparaenum}

\paragraph*{Overall Results}
The results we obtain for the different modalities, with and without the
 use of either CRF or sliding-window block sequence model, are shown in Table \ref{tab:crf_results}.
The following lessons can be drawn from these results:
\begin{asparaenum}
\item This is a hard task, as the best performance
  reached is 88\% for accuracy and 87\% for mean~$\textrm{F}_1$. Indeed,
  it can be hard even to a human to determine whether a block is part
  of a proof or theorem environment, especially in the middle of it, so
  it is unsurprising that we cannot reach near-perfect results.
\item Looking at unimodal models: the font-based model performs rather poorly,
  though still beating (at least in terms of mean $\textrm{F}_1$) the
  three baselines; the text-based model is the best
  performing one, suggesting that textual clues impact more than visual
  ones for this task.
\item The multimodal model outperforms every unimodal model, though the
  margin with the text model is somewhat low.
\item Including the Sequential model (both CRF, SW transformer, or HAT)
  greatly increases the performance of every unimodal or multimodal
  model, by 5 to 10 points of accuracy or mean $\textrm{F}_1$. The
  importance of the use of an approach modeling block sequences is thus
  clear. Long-distance dependencies captured by HATs do not seem to
  matter.
\end{asparaenum}

\paragraph*{Multimodal approach.}
To look more in detail at the impact of the choice of multimodal fusion
strategies, we report the performance of a variety of them in
Table~\ref{tab:model_comparison} with the cross-modal attention
technique described in Section~\ref{sec:beyond-unimodal}
highlighted in bold. We note that results of most multimodal approaches
are actually quite close to each other, which hints at the robustness of our
observation that adding a multimodal model on top of our three unimodal
models improves in all cases the performance on our classification task.

\paragraph*{Individual modalities.}
\label{seq:text_exp}

We now discuss the performance of different unimodal models.

\begin{table}
\centering\small
\caption{Performance comparison of text models}
\vspace{-1em}
\hspace*{-1em}\begin{tabular}{>{\bfseries} l@{}ccccccc}
\toprule
\bfseries & \bfseries & \bfseries Inf. time & \bfseries
Accuracy & \bfseries Mean $\textrm{F}_1$ & \bfseries \\
\bfseries Model & \bfseries \#Batches & \bfseries (ms/step)& \bfseries
(\%) & \bfseries (\%) & \bfseries \#Params \\
\midrule
Dummy& --- & ---  & 59.41 & 24.85 & ---  \\
\textsf{RoBERTa} base  & 20 & 23 & 76.61 & 71.66 & 124M \\
Pretrained & 20 & 23 & 76.45 & 72.33 & 124M \\
\textsf{SciBERT} base  & 20 & 23 & \bfseries 76.89 & \bfseries 73.00 & 110M \\
\bottomrule
\end{tabular}%
\label{tab:my-table_3}
\end{table}

\begin{toappendix}
\subsection{Experimental Results}

We show how different models (across different modalities) scale
performance with the increasing data in Figures~\ref{fig:nlp_train},
\ref{fig:vis_train}, and \ref{fig:font_train}, respectively for the text,
vision, and font models.

\begin{figure}
\centering  \includesvg[width=.5\linewidth]{images/nlp_train_1.svg}
    \caption{Accuracy of language models on fine-tuning task with respect
    to number of batches}
    \label{fig:nlp_train}
\end{figure}

\begin{figure}
\centering  \includesvg[width=.5\linewidth]{images/inc_vis.svg}
    \caption{Accuracy of Vision models on fine-tuning task with respect
    to number of batches}
    \label{fig:vis_train}
\end{figure}

\begin{figure}
\centering  \includesvg[width=.5\linewidth]{images/inc_font.svg}
    \caption{Accuracy of Font models on fine-tuning task with respect
    to number of batches}
    \label{fig:font_train}
\end{figure}
\clearpage
\end{toappendix}

\begin{table}
\centering
\small
\caption{Samples to target accuracy for text
models}
\vspace{-1em}
\hspace*{-1.5em}
\begin{tabular}{>{\bfseries}llcc}
\toprule
Model & \bfseries Data size & \bfseries Samples to 65\% &\bfseries  Samples to 70\%\\
\midrule
\textbf{RoBERTa} base & 160~GB & 41\,472 & 186\,496 \\
Pretrained & 11~GB, 197k papers & 39\,552 & 141\,632  \\
\textsf{SciBERT} base & 1.14M papers & 36\,928 & 91\,200 \\
\bottomrule
\end{tabular}
\label{table:few-shot}
\end{table}

To measure the performance of various language models (our language model
pretrained on our corpus, \textsf{RoBERTa}, and \text{SciBERT}),
we evaluate their
accuracy as shown in Table~\ref{tab:my-table_3} on the
validation dataset. All three have similar numbers of parameters
(obviously the same for our pretrained and the base version of
\textsf{RoBERTa}), have similar inference time, and reach similar levels of accuracy ($76.45\%$
to $76.89\%$) and mean $\textrm{F}_1$ (71.66\% to 73.00\%) and converge after training on 20 batches. SciBert does
have slightly higher performance.
Table~\ref{tab:my-table_3} shows another side of the picture: to reach a
target level of accuracy (say, 65\% or 70\%), our pretrained model
needs much fewer fine-tuning data
than the \textsf{RoBERTa} model (trained on a corpus of 15
times more text data). Although \textsf{SciBERT} performs better than our
pretrained model on this metric as well, note that it
has been trained on 5.5 times more scientific
papers than our pretrained model.


\begin{table*}
\centering
\caption{Performance comparison of vision models}
\vspace{-1em}
\hspace*{-2.5em}\begin{tabular}{>{\bfseries}l@{}cccccccc}
\toprule
\bfseries & \bfseries & \bfseries Inf. time & \bfseries
Accuracy & \bfseries Mean $\textrm{F}_1$ & \bfseries \\
\bfseries Model & \bfseries \#Batches & \bfseries (ms/step)& \bfseries
(\%) & \bfseries (\%) & \bfseries \#Params &\\
\midrule
Dummy& - & - & 59.41 & 24.85 & - \\
\textsf{EfficientNetB0}& 5 & 29 & 65.27 & 46.00 & 6.9M \\
\textsf{EfficientNetB0\_max }& 5 & 35 & 58.22 & 25.00 & 4.0M \\
\textsf{EfficientNetB0\_avg }& 5 & 34 & 62.93 & 39.66 & 4.0M \\
\textsf{EfficientNetB4\_avg }& 5 & 61 & 65.87 & 47.33 & 17.6M \\
\textsf{EfficientNetB7\_avg }& 5 & 145& 61.22 & 42.33 & 64.1M \\
\textsf{EfficientNetV2s\_avg }& 5 &70 & 59.41 & 25.00 & 20.3M \\
\textsf{EfficientNetV2m\_avg }& 5 &94 & 64.02 & 42.66 & 53.2M \\
\textsf{EfficientNetB4\_avg }& 9 & 88 & 68.47 & 54.33 & 17.6M \\
\textsf{EfficientNetV2s\_avg }& 9 &71 & 59.81 & 27.00 & 20.3M \\
\textsf{EfficientNetV2m\_avg }& 9 &92 & \bfseries 69.44 &
\bfseries 60.33 & 53.2M \\
\bottomrule
\end{tabular}%
\label{tab:my-table}
\end{table*}

\begin{table*}
\centering
\caption{Performance comparison of font models}
\vspace{-1em}
\begin{tabular}{>{\bfseries}lccccccc}
\toprule
\bfseries & \bfseries & \bfseries Inf. time & \bfseries
Accuracy & \bfseries Mean $\textrm{F}_1$ & \bfseries \\
\bfseries Model & \bfseries \#Batches & \bfseries (ms/step)& \bfseries
(\%) & \bfseries (\%) & \bfseries \#Params \\
\midrule
\label{table:my-table_1}
Dummy& - & -  & 59.41 & 24.85 & -  \\
LSTM (128) & 11 & 14 & 64.93 & 45.48 & 1.72M \\
GRU (128) & 11 & 14 & 60.59 & 42.71 & 1.72M \\
BiLSTM (128) & 11 & 26 & 64.71 & 45.66 & 1.82M \\
\bottomrule
\end{tabular}
\end{table*}

\begin{table*}
    \centering
    \small
    \caption{Class-wise precision and recall scores for best unimodal,
    multimodal, and sequential models}
    \vspace{-1em}
    \begin{tabular}{rcccccccccccc}
        \toprule
        & \multicolumn{2}{c}{\emph{Font}} &
        \multicolumn{2}{c}{\emph{Vision}} &
        \multicolumn{2}{c}{\emph{Text}} &
        \multicolumn{2}{c}{\emph{Multimodal}} &
        \multicolumn{2}{c}{\emph{Sequence CRF}} & \multicolumn{2}{c}{\emph{Sequence Transformers}} \\
        \cmidrule(lr){2-3}\cmidrule(lr){4-5}\cmidrule(lr){6-7}
        \cmidrule(lr){8-9}\cmidrule(lr){10-11}\cmidrule(lr){12-13}
        & \bfseries Precision & \bfseries Recall &
        \bfseries Precision & \bfseries Recall & \bfseries Precision &
        \bfseries Recall
        & \bfseries Precision & \bfseries Recall &
        \bfseries Precision & \bfseries Recall & \bfseries Precision &
        \bfseries Recall \\
        \cmidrule(lr){2-3}\cmidrule(lr){4-5}\cmidrule(lr){6-7}
        \cmidrule(lr){8-9}\cmidrule(lr){10-11}\cmidrule(lr){12-13}
        \bfseries Basic & 0.6534 & 0.9750 & 0.6902 & 0.9119 & 0.7963 & 0.8539
        & 0.7953 & 0.8863 & 0.8538 & 0.8993 & 0.9047 & 0.8970 \\
        \bfseries Theorem & 0.8657 & 0.3770 & 0.7778 & 0.6158 & 0.7498 & 0.6038
        & 0.8561 & 0.6845 & 0.8569 & 0.8019 & 0.8768 & 0.9030 \\
        \bfseries Proof & 0.5039 & 0.0375 & 0.6086 & 0.2223 & 0.6860 & 0.6717
        & 0.7129 & 0.6184 & 0.8022 & 0.7570 & 0.8157 & 0.8376 \\
        \bottomrule
    \end{tabular}
    \label{tab:unimodal_precision_recall}
\end{table*}

The performance of a wide variety of vision-based models is displayed in
Table~\ref{tab:my-table}. For the simplest model, we experiment with
different forms of pooling for the last convolutional layer: none, max
pooling, or average pooling. We see that no pooling yields performance,
on a small model that is quite close and comparable to much larger models.
We also notice that average pooling works quite well in most cases, while also cutting the number of parameters to nearly a third.

For font sequence information, in addition to our model formed of an LSTM with 128 cells, and in order
to investigate potential further gains, we try switching LSTM cells to
GRU~\cite{chung2014empirical}; and using a Bidirectional LSTM to capture sequential information across both forward and backward axis.
Our results from Table~\ref{table:my-table_1} indicate that the
bidirectional component in fonts alone does not have a huge impact in
deciding the label of the blocks, even if modest gains are observed.

We finally show in Table \ref{tab:unimodal_precision_recall} a partial classification report, for the best model
in each class.

\begin{toappendix}
\begin{figure}
\frame{\includegraphics[width=\linewidth]{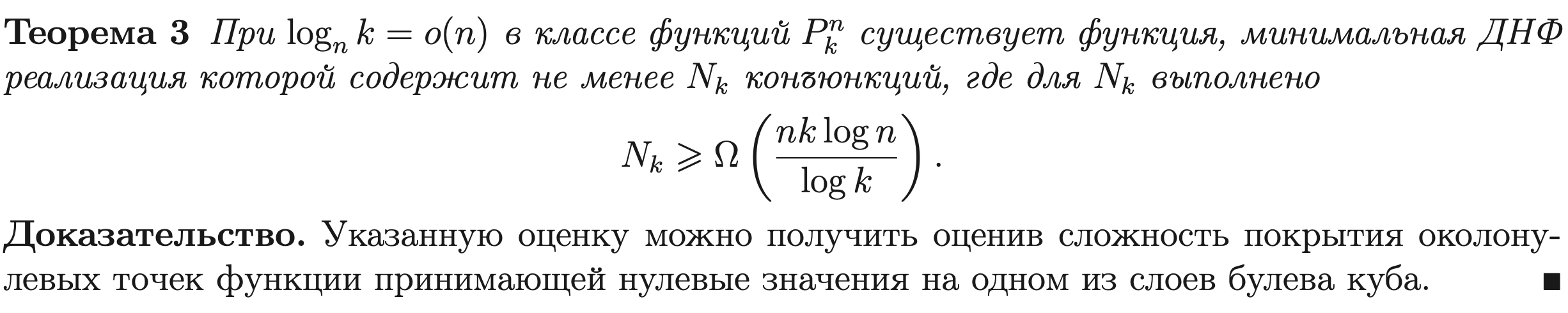}}
\caption{Excerpt from an example Russian-language paper~\cite{granin2015average}
with high performance of font model}
\label{fig:russian_proof}
\end{figure}
Despite the low score obtained by the font models, they can
still be of use in certain situations. For example, anecdotally,
Figure~\ref{fig:russian_proof} shows an example of a Russian-language
article
whose blocks are correctly classified by the font-based model, while the
text model is not able to use any clues as it was trained on
English text.

\end{toappendix}


\begin{toappendix}
    \subsection{Hyper parameter tuning of SW Transformer}
    In this section, we present our investigation into the hyperparameters for our sliding window (SW) transformer model. Initially, we examined a configuration akin to the original BERT architecture, featuring 16 attention heads, and varied the maximum sequence length (maxlen)\footnote{It is important to note that if a document exceeds the specified maxlen, it is divided into two separate segments, each treated as an independent document with padding added to maintain dimensional consistency.}. Our findings, documented in Table \ref{tab:heads_performance}, indicate that a larger maxlen detrimentally affects model performance. Consequently, we established a maxlen of 256 and proceeded to experiment with varying the number of attention heads, as detailed in Table \ref{tab:heads_performance}. Further experimentation was conducted with even smaller maxlen values while maintaining 20 attention heads, a configuration derived from Table \ref{tab:model_performance}.

    Subsequent investigations focused on the effect of varying the number of hidden units in the feedforward network and the number of encoder stacks, with results presented in Tables \ref{tab:ff_dim_performance} and \ref{tab:encoder_block_performance}, respectively. Based on the insights gained from Table \ref{tab:ff_dim_performance}, we developed a more parameter-efficient encoder block configuration, employing 1.5 times the number of feedforward units, which demonstrated superior performance and efficiency compared to the 4x setting recommended in the original BERT paper, as shown in Table \ref{tab:transformer_comparison}.

    Additionally, we incorporated a sliding window mechanism into the model, as outlined in Table \ref{tab:architecture_optimization}, and conducted a detailed exploration of window sizes, documented in Table \ref{tab:sliding_attention_mechanism}. Upon identifying an optimal sliding window size, we increased the number of training epochs to offset the impact data size when using smaller maxlen inputs, with the outcomes reported in Table \ref{tab:lower_maxlen_impact}.

\begin{table}[ht]
\centering
\caption{Impact of large Max Length on Transformer Model}
\label{tab:model_performance}
\begin{tabular}{lcccc}
\toprule
Max Len & Train loss & \bfseries Accuracy (\%) & \bfseries Mean $\textrm{F}_1$ (\%)  \\
\midrule
1024 & 0.4984 & 80.37 & 77.06 \\
512 & 0.4953 & 80.46 & 77.39 \\
256 & 0.4565 & 80.65 & 78.64  \\
128 & 0.4843 & 78.79 & 76.16  \\
\bottomrule
\end{tabular}
\end{table}

\begin{table}[ht]
\centering
\caption{Impact of Attention Heads (with maxlen = 256) based on results from table \ref{tab:model_performance}}
\label{tab:heads_performance}
\begin{tabular}{lcccc}
\toprule
Heads & Train Loss & \bfseries Accuracy (\%) & \bfseries Mean $\textrm{F}_1$ (\%) & \\
\midrule
8  & 0.5014 & 80.80 & 78.11 \\
12 & 0.5007 & 80.83 & 78.06 \\
16 & 0.4986 & 80.86 & 78.13 \\
20 & 0.4992 & 80.84 & 78.15 \\
\bottomrule
\end{tabular}
\end{table}

\begin{table}[ht]
\centering
\caption{Impact of small Max Length on Transformer Model (with heads=20) based on results from table \ref{tab:heads_performance}}
\label{tab:max_len_performance}
\begin{tabular}{lccccc}
\toprule
Max Len & Train Loss & \bfseries Accuracy (\%) & \bfseries Mean $\textrm{F}_1$ (\%) & \\
\midrule
128 & 0.4846 & 81.20 & 78.61  \\
64 & 0.4609 & 82.03 & 79.48 \\
32 & 0.4307 & 85.46 & 83.90  \\
16 & 0.3890 & 85.17 & 85.01  \\
8 & 0.3970 & 83.16 & 81.15  \\
\bottomrule
\end{tabular}
\end{table}

\begin{table}[ht]
\centering
\caption{Impact of different ff-dim multipliers on Transformer Model (with maxlen=16, heads=20) based on results from table \ref{tab:max_len_performance}, \ref{tab:heads_performance})}
\label{tab:ff_dim_performance}
\begin{tabular}{lcccccc}
\toprule
ff-dim & Train Loss &  \bfseries \#Params & \bfseries Accuracy (\%) &  \bfseries Mean $\textrm{F}_1$ (\%) & \\
\midrule
6 times & 0.3905 & 26.90M & 85.03 & 84.88  \\
4 times & 0.3890 & 20.34M & 85.17 & 85.01 \\
2 times & 0.3888 & 13.79M & 85.24 & 84.90 \\
1 times & 0.3906 & 10.51M & 85.46 & 84.93  \\
0.5 times & 0.3967 & 8.87M & 84.81 & 84.51  \\
0.25 times & 0.3960 & 8.05M & 85.24 & 84.53  \\
\bottomrule
\end{tabular}
\end{table}

\begin{table}[ht]
\centering
\caption{Impact of Encoder Blocks (with maxlen=16, heads=20, \(ff\)-dim=\(1\times\) on Transformer model based on results from table \ref{tab:ff_dim_performance}, \ref{tab:heads_performance} , \ref{tab:max_len_performance})}
\label{tab:encoder_block_performance}
\begin{tabular}{lccccc}
\toprule
Encoders & Train Loss & \bfseries \#Params & \bfseries Accuracy (\%) &  \bfseries Mean $\textrm{F}_1$ (\%) \\
\midrule
1  & 0.3906 & 10.51M & 85.46 & 84.93  \\
2 & 0.3890 & 20.35M & 84.91 & 84.67 \\
\bottomrule
\end{tabular}
\end{table}

\begin{table}[ht]
\centering
\caption{Comparison of Bert like and Efficient Transformer Models (with maxlen=16, heads=20, encoders=1)}
\label{tab:transformer_comparison}
\begin{tabular}{lcccc}
\toprule
Model & Train Loss & \bfseries \#Params & \bfseries Accuracy (\%) &  \bfseries Mean $\textrm{F}_1$ (\%) \\
\midrule
BERT like \(ff\)-dim=\(4\times\) & 0.3890 & 20.34M & 85.17 & 85.01 \\
Efficient \(ff\)-dim=\(1.5\times\) & 0.3864 & 12.15M & 85.63 & 85.25 \\
\bottomrule
\end{tabular}
\end{table}

\begin{table}[ht]
\centering
\caption{Impact of SW mechanism of (window size=16) applied to encoder architecture found in table \ref{tab:transformer_comparison}}
\label{tab:architecture_optimization}
\begin{tabular}{ccc}
\toprule
Max Len & \bfseries Accuracy (\%) & \bfseries Mean $\textrm{F}_1$ (\%) \\
\midrule
1024 & 82.23 & 80.82 \\
512 & 82.84 & 80.68 \\
256 & 83.04 & 80.62 \\
128 & 84.45 & 83.22 \\
64 & 86.26 & 85.51 \\
32  & 86.59 & 85.97 \\
16 & 86.33 & 85.65 \\
\bottomrule
\end{tabular}
\end{table}

\begin{table}[ht]
\centering
\caption{Impact of Window Size (Maxlen =32) on Transformer model based on the results from table \ref{tab:architecture_optimization})}
\label{tab:sliding_attention_mechanism}
\begin{tabular}{lcccc}
\toprule
\bfseries Sliding Window & Train Loss & \bfseries Accuracy (\%) & \bfseries Mean $\textrm{F}_1$ (\%) \\
\midrule
32 & 0.3728  & 85.39 & 84.537 \\
16 & 0.3496  & 86.78 & 86.079 \\
8 & 0.3639 & 86.15 & 85.438 \\
4 & 0.3903 & 84.55 & 83.361 \\
\bottomrule
\end{tabular}
\end{table}

\begin{table}[ht]
\centering
\caption{Increasing the number of Epochs to counter smaller maxlen (SW=16) based on the results from table \ref{tab:sliding_attention_mechanism}}
\label{tab:lower_maxlen_impact}
\begin{tabular}{lccccc}
\toprule
\textbf{Maxlen} & \textbf{Validation Samples} & \textbf{Epochs} & \textbf{Train Loss} & \textbf{Accuracy (\%)} & \bfseries Mean $\textrm{F}_1$ (\%) \\
\midrule
32 & 18K & 1 & 0.3465 & 86.80 & 86.21 \\
64 & 10K & 2 & 0.3415 & 86.70 & 86.05 \\
128 & 6K & 4 & 0.3352 & 87.30 & 86.76 \\
256 & 5K & 8 & 0.3157 & 87.33 & 86.99 \\
512 & 4K & 16 & 0.2771 & 87.52 & 86.58 \\
1024 & 4K & 32 & 0.2726 & 87.81 & 87.18 \\
2048 & 4K & 32 & 0.2692 & 86.58 & 85.41 \\
\bottomrule
\end{tabular}
\end{table}
\end{toappendix}

\begin{toappendix}
  \clearpage
  \subsection{Interpretability}
Though our models are not directly interpretable, we can get some
post-hoc explanations of their performance: for the text modality, we can
visualize~\cite{vig-2019-multiscale} the attention heads,
see Figure~\ref{fig:nlp_vis} where we visualize the last
layer of the language model to see what the model focuses on.
Informal
experiments suggest, unsurprisingly, that tokens indicative of
these environments (such as the ``Theorem'' or ``Proof'' tokens, or
numberings) are given a strong weight in determining the label of the
paragraphs. For the vision model, we can use the
Grad-CAM~\cite{selvaraju2017grad} visualization, see Figure
\ref{fig:grad_vis}, which indeed shows
that some layout-based information is
captured by the model.
\end{toappendix}

\bigskip

\section{Conclusion}
Summarizing the results obtained in the previous section, we put forward our multimodal model with
block-sequential sliding-window transformer model as a state-of-the-art candidate for
identification of theorems and proofs for scientific articles. The level
of accuracy and mean $\textrm{F}_1$ reached, if not perfect, is acceptable for
automatic processing of articles and construction of a knowledge base of
theorems, which may need to be further manually cleaned and curated.
Text stands out as the most effective single modality for our analysis,
surpassing vision and font sequence in performance, though the latter boasts the highest efficiency and minimal parameter usage, approximately 70 times less than our language model.


An important distinctive feature of our research is that we focus on
analyzing scientific documents spanning multiple pages, unlike typical
document AI methods designed for single-page documents (e.g., receipts,
bills) and simpler tasks (e.g., total bill calculation, document type
classification). 

Here are some advantages of our approach:
\begin{asparaitem}
    \item \textbf{\textit{Processing entire PDFs while capturing sequential dependencies:}} Our model processes entire PDFs, generating labels for each paragraph in a single forward pass and capturing sequential dependencies across pages. For reference, our 198M model can process an entire PDF at once, whereas LayoutLMv3 \cite{huang2022layoutlmv3} (368M) and Nougat (250M) \cite{blecher2023nougat} can only process one page at a time.
    \item \textbf{\textit{Modular and multimodal:}} Our approach is both
      multimodal and modular. We demonstrate this by integrating a custom
      pre-trained Roberta model, which will allow us to switch to
      different text backbones to extend this work (e.g., LLAMA \cite{touvron2023llama}, SPECTER \cite{DBLP:conf/acl/CohanFBDW20},
      ORCA-MATH \cite{DBLP:journals/corr/abs-2402-14830}) without redesigning the entire architecture as for
      future versions of the LayoutLM family. This flexibility mirrors
      the CLIP \cite{DBLP:conf/icml/RadfordKHRGASAM21} model's approach (see Table 10 of the CLIP paper) and is not possible with other models like LayoutLM.
    \item \textbf{\textit{Scalability and speed:}} Unlike LayoutLM models
      that require OCR (adding to inference time, see $\alpha$ parameter in table 1
      of Donut paper that factors the OCR time when comparing to LayoutLM). While Nougat and Donut are OCR-free and
      faster, they are still slower than our approach, to be used in a
      real-world setting to help researchers. For reference, Nougat takes
      19.6 seconds for 6 pages on an Nvidia A10G, whereas Grobid
      processes 10.6 PDFs/sec on a CPU (see limitation section of Nougat
      paper). This efficiency is crucial to evaluate our dataset of 200k
      papers, which would take several months. We also cut down the number of parameters by switching the vanilla transformer encoder block to a more efficient sliding window encoder block that reduces the number of parameters leading to a reduced inference time and memory usage.
    \item \textbf{\textit{Alternate and lighter models:}} Unlike typical model sizes
      offering small and large variants of the same model (in terms of parameter count)
      models, we provide models at different modality levels. This allows
      for the integration of extremely lighter alternatives, such as a 2M parameter
      CRF model trained on fonts, which achieves 71\% accuracy for an extremely low-resource setting, see table \ref{tab:unimodal_precision_recall} for per class modality performance.
\end{asparaitem}

One limitation of our work is that we have only trained and evaluated our model on English-language
mathematical articles. Though they represent a significant portion of the
mathematical literature, articles in other languages do exist and one would
need to check whether the approach proposed extends to, say, Russian- or
Arabic-language articles. This can be extended by removing the English language filter in the preprocessing step.

Our approach is  a first building block towards building a knowledge base of
theorems from the raw PDFs, but further research is required before
one will be able to provide such an application.

\balance
\section{Ethical Considerations}
We do not envision any major ethical concerns from the development
of machine learning models to extract mathematical statements and proofs
from mathematical articles.

As for all imperfect methods, results should not be blindly used
in settings where perfect accuracy is required.

The training of machine learning models in general, and deep learning
models in particular, requires a significant amount of computation
time and energy consumption, which contributes to the production of
greenhouse gases and global warming. We attempted to somewhat mitigate
this by focusing on models with good performance but
reasonable numbers of parameters. We also note that the supercomputer
used for the computation is powered with low-carbon electricity, and that
residual heat produced by the computing power is used as part of an urban heat
distribution network.

Finally, note that though we used publicly available research articles
from arXiv, acquired respecting arXiv's terms and conditions,
redistribution of the dataset is not allowed by the arXiv licensing
agreement (except for the few papers that are explicitly marked with a
Creative Commons license). Distribution of the model learned from this
publicly available dataset is somewhat of a grey legal area (as is the
release of all machine learning models trained on publicly available
data without a specific license of use, such as most openly available
large language models). To allow reproducibility while respecting
licensing terms, we provide at \url{https://github.com/mv96/mm_extraction}:
full instructions on how to rebuild the same dataset by
retrieving data from arXiv; all ground truth annotations (label of each
paragraph block of each article in the dataset); trained models;
full code to train them, released as free software.

\section*{Acknowledgments}
\label{sec:acknowledgements}

This work was funded in part by the French government under management of Agence Nationale
de la Recherche as part of the “Investissements
d’avenir” program, reference ANR-19-P3IA-0001
(PRAIRIE 3IA Institute).
This work was also made possible through HPC resources of IDRIS
granted under allocation 2020-AD011012097
made by GENCI (Jean Zay supercomputer).

\clearpage

\bibliographystyle{ACM-Reference-Format}
\bibliography{bibliography}
\end{document}